\documentclass[letterpaper]{article} 
\usepackage{aaai2027} 
\usepackage[hyphens]{url} 
\usepackage{graphicx} 
\def\UrlFont{\rm} 
\usepackage{natbib} 
\usepackage{caption} 
\usepackage{amsmath}
\usepackage{amssymb}
\usepackage{booktabs}
\usepackage{multirow}
\usepackage{pifont}
\usepackage{algorithm}
\usepackage{algorithmic}
\usepackage{dblfloatfix}
\usepackage{placeins}

\newcommand{\cmark}{\ding{51}}
\newcommand{\xmark}{\ding{55}}
\newcommand{\ms}[2]{#1{\scriptsize$\pm$#2}}

\title{DHRCL: Training Code LLMs with Dense Hierarchical Rewards and Curriculum Learning}
\author{
    Shuhang Wang,
    Ziming Li,
    Hui Cheng
}
\affiliations{
}

\begin{document}
\maketitle

\begin{abstract}
Reinforcement learning is a natural post-training paradigm for code-oriented large language models because generated programs can be evaluated through parsing, execution, unit tests, and structural analysis.
However, existing methods often rely on sparse outcome rewards or statically combine heterogeneous dense signals, even though syntax validity, executability, functional correctness, and structural organization describe different and progressively dependent programming capabilities.
We propose \textbf{DHRCL}, a reinforcement learning framework with Dense Hierarchical Rewards and Curriculum Learning.
DHRCL decomposes feedback into syntax validation, execution success, unit-test pass rate, and AST-based structural similarity, and organizes these signals through a three-stage Syntax, Execution, Pass \& Structural curriculum.
Stage duration is determined automatically from recent validation trends rather than manually specified capability thresholds.
We further introduce stage-aware probability-based token credit redistribution.
The mechanism follows a consolidation-to-refinement principle: it emphasizes established token patterns during syntax-oriented optimization, applies uniform propagation for non-local execution feedback, and allocates more credit or blame to less-established token decisions during final functional optimization.
Under a unified Qwen3-8B and KodCode protocol, the experiments compare DHRCL with binary, pass-rate, reward-model-based, and verifiable dense-reward baselines.
We further evaluate DHRCL across Qwen3-4B, Qwen3-8B, and Qwen3-14B backbones, showing that its advantage remains consistent as model capacity increases.
\end{abstract}

\section{Introduction}
Large language models (LLMs) have made code generation a central application of modern language models~\citep{codex,codellama}, while realistic benchmarks increasingly emphasize execution-based validation~\citep{bigcodebench,odex}.
Code is well suited to reinforcement learning (RL) because generated programs admit automatically verifiable feedback from parsing, execution, tests, and structural analysis.
Prior work exploits compiler and unit-test outcomes~\citep{coderl,ppocoder,rltf}, but sparse rewards provide little guidance when most samples fail, while static combinations of dense rewards ignore the prerequisite relations among syntax validity, executability, and functional correctness.

We propose \textbf{DHRCL}, a code-LLM RL framework with \textbf{D}ense \textbf{H}ierarchical \textbf{R}ewards and \textbf{C}urriculum \textbf{L}earning.
DHRCL combines syntax validation, execution success, unit-test pass rate, and AST-based structural similarity in a Syntax, Execution, Pass \& Structural curriculum.
Stage duration is selected automatically from held-out validation trends rather than manually chosen capability thresholds.
DHRCL also redistributes trajectory-level credit across tokens according to a consolidation-to-refinement principle: confidence-oriented weighting in the syntax stage, uniform weighting for non-local execution feedback, and uncertainty-oriented weighting in the final stage.

Our contributions are:
\begin{itemize}
    \item A three-stage hierarchical reward curriculum coordinating syntax, execution, functional, and structural feedback while retaining functional correctness as the primary objective.
    \item Trend-based automatic stage progression that adapts stage duration without fixed syntax or execution thresholds.
    \item Stage-aware token credit redistribution with controlled ablations across reward composition, curriculum scheduling, and token weighting.
\end{itemize}

\section{Related Work}

\subsection{RL-Based Optimization for Code LLMs}
RL has become a prominent post-training paradigm for aligning code LLMs with sequence-level functional correctness~\cite{survey01,survey02}.
Early policy-gradient methods such as REINFORCE~\cite{reinforce} enable optimization beyond token likelihood, and have been adapted to code generation through actor-critic and PPO-style frameworks.
CodeRL~\cite{coderl} uses execution feedback in an actor-critic setting, while PPOCoder~\cite{ppocoder}, RLTF~\cite{rltf}, and RLEF~\cite{RLEF} extend PPO~\citep{ppo} to compiler-guided, online, and iterative feedback scenarios.
Related preference- and group-based methods further improve scalable post-training: DPO~\cite{dpo} and Focused-DPO~\cite{focuseddpo} avoid explicit critic learning, while GRPO~\cite{dsmath,dsr1}, DAPO~\cite{dapo}, and GSPO~\cite{gspo} improve critic-free or sequence-level policy optimization.
These methods strengthen the RL optimization backbone, whereas our work focuses on how code-specific rewards are constructed and organized.

\subsection{Code-Specific Reward Design}
Reward design is central to RL-based code generation because programs can be verified through compilation, execution, tests, and structural analysis.
Existing rewards include verifiable execution or compiler feedback~\cite{coderl,ppocoder,rltf,RLEF}, structured or dense signals such as AST/DFG consistency, partial test success, and task-specific code similarity~\cite{rlhwgen,VeRPO,ReCode}, as well as learned reward models and process-level supervision~\cite{AceCoder,codeprm,Process-SupervisedRL}.
Recent preference- and test-construction methods further refine supervision through test-based preferences or adversarial test generation~\cite{PLUM,CodeDPO,Target-DPO,Code-A1}.
Overall, these studies enrich code rewards with verifiable, dense, local, or adaptive feedback, but heterogeneous objectives are usually combined statically.
This leaves stage-wise reward scheduling across syntax, execution, tests, and structure underexplored.

\subsection{Curriculum Learning for Code LLMs}
Curriculum learning organizes optimization from easier to harder learning units.
In code generation, prior work has explored curricula over completion subtasks, data complexity, code blocks, and test cases.
StepCoder~\cite{stepcoder} decomposes full program synthesis into code-completion subtasks with fine-grained optimization; data-level curricula order training samples by code complexity to improve small code LMs~\cite{small}; AST-guided block splitting improves fill-in-the-middle DPO alignment~\cite{Alignment}; and TAROT~\cite{tarot} constructs tiered test suites with capability-adaptive curriculum policies.
These methods mainly schedule training units rather than heterogeneous reward objectives.
In contrast, DHRCL organizes syntax, execution, functional, and structural rewards into a stage-aware curriculum and further propagates them with stage-aware token credit redistribution.

\section{Method}

\begin{figure*}[t]
\centering
\includegraphics[width=\textwidth]{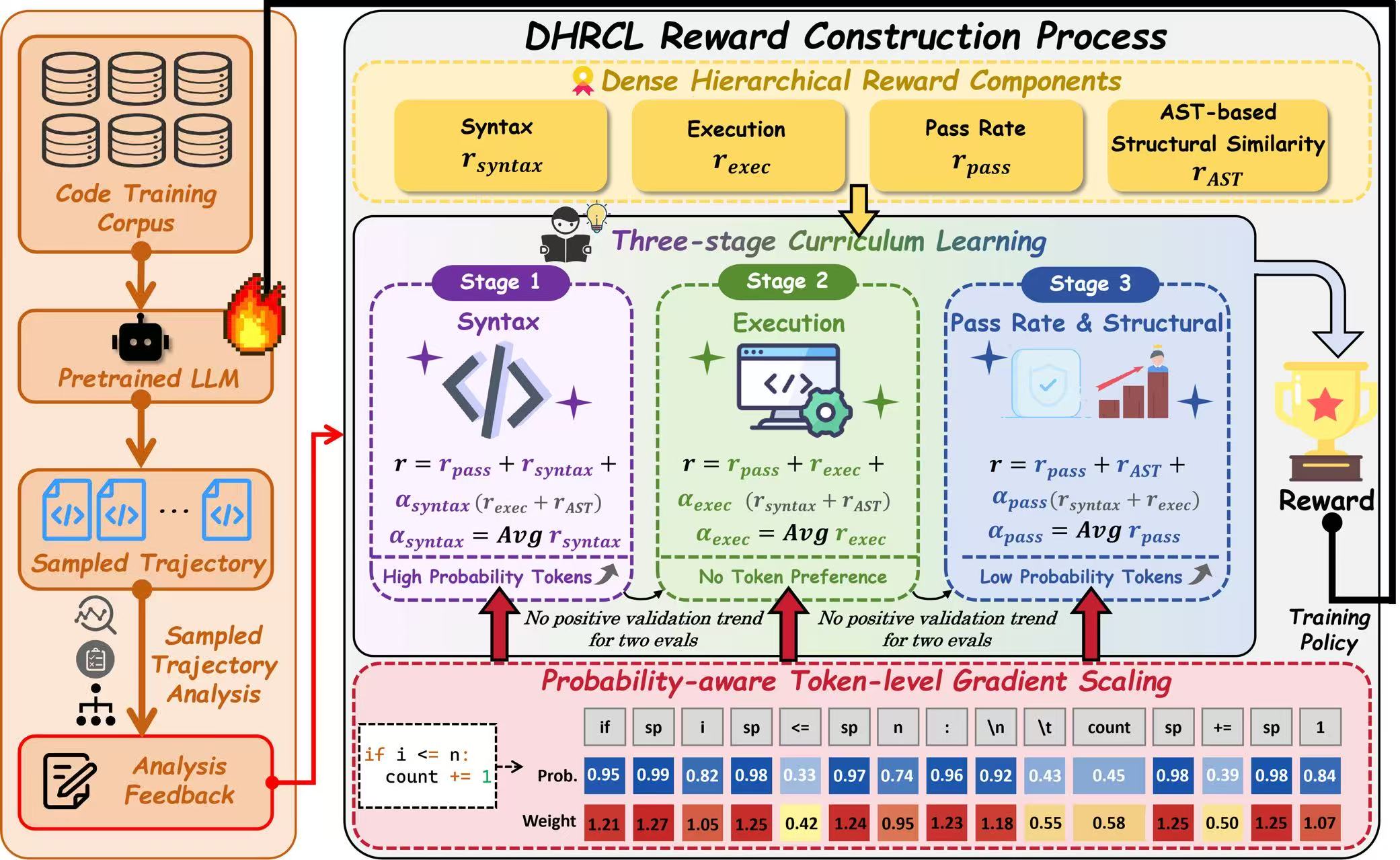}
\caption{Overall pipeline of DHRCL.}
\label{fig:pipeline}
\end{figure*}

We propose \textbf{DHRCL}, a reinforcement learning framework for code-oriented LLM post-training.
As shown in Figure~\ref{fig:pipeline}, DHRCL contains two coupled components.
First, a dense hierarchical reward decomposes code feedback into syntax validity, execution validity, partial functional correctness, and AST-based structural alignment.
Second, a stage-wise curriculum changes the primary optimization emphasis according to the policy's current bottleneck, while probability-aware token credit redistribution controls how each trajectory-level update is allocated across generated tokens.
The predefined capability order is Syntax, Execution, Pass \& Structural, whereas each stage duration is determined automatically from held-out validation trends.

\subsection{Dense Hierarchical Reward}

Outcome-level rewards such as binary test success or aggregate pass rate often cannot distinguish locally malformed code, runtime failures, and partially correct executable programs.
DHRCL therefore combines four complementary reward components.
The first three directly describe program validity and functional behavior.
The fourth is an auxiliary, reference-conditioned structural signal; it is not treated as semantic equivalence or as a complete measure of code quality.

\subsubsection{Syntax Validation Rate}

We use a parser-based syntax reward to provide partial feedback for locally well-formed code.
Given a generated program $y$, an error-tolerant parser extracts syntactic units $\mathcal{B}(y)=\{b_i\}_{i=1}^{m}$.
A unit is valid if its subtree contains no parser error node, and the reward is
\[
r_{\mathrm{syntax}}(y)=
\frac{\sum_{i=1}^{m} w_i v_i}
{\sum_{i=1}^{m} w_i},
\]
where $v_i$ is the validity indicator and $w_i$ is the token length.
We set $r_{\mathrm{syntax}}(y)=0$ if no unit is extracted.
Implementation details are provided in the Supplementary Document, Section ``Implementation Details of Syntax Reward.''

\subsubsection{Execution Success}

While syntax validity ensures that a program can be parsed at least locally, it does not guarantee successful execution.
Generated code may still trigger runtime errors, exceed time limits, or cause memory-related failures.
To capture runtime validity, we introduce an execution success reward:
\[
r_{\text{exec}} =
\mathbb{I}(\text{execution success}),
\]
where $\mathbb{I}(\cdot)$ denotes the indicator function.
The reward is 1 if the generated program executes successfully under the evaluation environment, and 0 otherwise.
This component encourages the model to move beyond syntactic well-formedness and learn programs with valid runtime behavior.

\subsubsection{Pass Rate}

The final goal of code generation is to solve the programming problem, which is typically evaluated by unit tests.
Instead of using a binary all-or-nothing reward, we use the fraction of passed test cases as a dense functional correctness signal:
\[
r_{\text{pass}} =
\frac{n_{\text{pass}}}
{n_{\text{test}}},
\]
where $n_{\text{pass}}$ is the number of passed test cases and $n_{\text{test}}$ is the total number of test cases.
This reward differentiates partially correct solutions and provides a smoother optimization signal than binary test success.

\subsubsection{AST-Based Structural Similarity}

Unit-test feedback evaluates functional behavior but does not distinguish the structural organization of generated programs.
We therefore introduce an auxiliary reference-conditioned AST reward.
For a generated program $y$ and a verified reference solution $y^{\mathrm{ref}}$, let $\mathcal{N}(\mathrm{AST}(\cdot))$ denote the structural normalization used by our implementation.
For programs that can be strictly parsed, we compute
\[
r_{\mathrm{AST}}(y)
=
\mathrm{Sim}_{\mathrm{AST}}
\left(
\mathcal{N}(\mathrm{AST}(y)),
\mathcal{N}(\mathrm{AST}(y^{\mathrm{ref}}))
\right),
\]
where $\mathrm{Sim}_{\mathrm{AST}}\in[0,1]$ is the implemented AST-similarity function.
If the generated program cannot be strictly parsed, we set $r_{\mathrm{AST}}(y)=0$.
Implementation details of the AST structural reward are provided in the Supplementary Document, Section ``Implementation Details of the AST Structural Reward.''

AST similarity is not interpreted as semantic equivalence, maintainability, or computational efficiency.
Instead, it supplies a weak structural prior derived from a verified implementation.
Functional correctness remains directly supervised by the unit-test pass-rate reward, while the AST signal encourages exploration of structurally coherent implementation patterns.
The reward-component ablation compares the full formulation against an otherwise identical variant without $r_{\mathrm{AST}}$, thereby isolating its contribution to Pass@1, structural similarity, and the observed characteristics of accepted outputs.

\subsection{Curriculum Learning for Code Generation}

Although the above rewards provide dense supervision, combining them with fixed weights ignores prerequisite relations among programming capabilities.
DHRCL uses three consecutive stages: Syntax, Execution, and Pass \& Structural.
The curriculum changes the primary optimization emphasis rather than completely enabling or disabling a reward.
In particular, $r_{\mathrm{pass}}$ remains active throughout training because functional correctness is always the final task objective, while previously introduced validity signals remain auxiliary components in later stages.

\subsubsection{Syntax Stage}

For the $i$-th sampled trajectory, the Syntax-stage reward is
\[
r_i =
r_{\mathrm{pass},i}
+
r_{\mathrm{syntax},i}
+
\alpha_{\mathrm{syntax}} r_{\mathrm{exec},i}
+
\alpha_{\mathrm{syntax}} r_{\mathrm{AST},i},
\]
where
\[
\alpha_{\mathrm{syntax}}
=
\frac{1}{n}\sum_{i=1}^{n}r_{\mathrm{syntax},i}.
\]
Here, $n$ is the number of trajectories in the rollout batch.
The batch-level coefficient acts as a competence-dependent gate.
When syntax validity is weak, downstream execution and structural signals receive less weight; as syntax validity improves, they are introduced more strongly.
The pass-rate reward is retained without attenuation.

\subsubsection{Execution Stage}

After syntax-oriented optimization, the primary emphasis shifts to executable behavior:
\[
r_i =
r_{\mathrm{pass},i}
+
r_{\mathrm{exec},i}
+
\alpha_{\mathrm{exec}} r_{\mathrm{syntax},i}
+
\alpha_{\mathrm{exec}} r_{\mathrm{AST},i},
\]
with
\[
\alpha_{\mathrm{exec}}
=
\frac{1}{n}\sum_{i=1}^{n}r_{\mathrm{exec},i}.
\]
Runtime validity often depends on interactions among definitions, control flow, data types, API calls, and memory or time constraints.
This stage therefore prioritizes whole-program execution feedback while retaining syntax and structural signals as auxiliary terms.

\subsubsection{Pass \& Structural Stage}

In the final stage, the primary objective combines functional correctness with the auxiliary structural prior:
\[
r_i =
r_{\mathrm{pass},i}
+
r_{\mathrm{AST},i}
+
\alpha_{\mathrm{pass}} r_{\mathrm{syntax},i}
+
\alpha_{\mathrm{pass}} r_{\mathrm{exec},i},
\]
where
\[
\alpha_{\mathrm{pass}}
=
\frac{1}{n}\sum_{i=1}^{n}r_{\mathrm{pass},i}.
\]
The pass-rate term is the direct functional objective.
The AST term is an auxiliary reference-conditioned signal and does not replace execution-based correctness.
Final checkpoint selection is based on validation Pass@1, with AST similarity used only as a tie-breaker between checkpoints with identical validation Pass@1.

\subsubsection{Trend-Based Automatic Stage Progression}

Fixed absolute thresholds can immediately advance a strong initialization or indefinitely retain a weak one, even when both policies exhibit meaningful learning progress.
DHRCL instead determines stage transitions from the recent trend of the primary held-out validation metric.
Let $m_s^{(k)}$ be the primary validation metric of stage $s$ at the $k$-th evaluation, where $k$ indexes validation observations:
\[
m_s^{(k)}
=
\begin{cases}
r_{\mathrm{syntax}}^{\mathrm{val},(k)}, & s=\text{Syntax},\\
r_{\mathrm{exec}}^{\mathrm{val},(k)}, & s=\text{Execution},\\
\mathrm{Pass@1}_{\mathrm{val}}^{(k)}, & s=\text{Pass \& Structural}.
\end{cases}
\]
After at least 12 validation observations have been collected in the current stage, we use the most recent eight observations,
\[
\mathcal{H}_s^{(k)}
=
\left\{
m_s^{(k-7)},\ldots,m_s^{(k)}
\right\},
\]
and fit the linear model
\[
m_s^{(t)}=a_s+b_s t+\epsilon_t.
\]
We perform a one-sided test of
\[
H_0:b_s\leq0
\qquad\text{against}\qquad
H_1:b_s>0.
\]
The policy remains in the current stage while the recent window supports a positive trend.
A transition is triggered only after the positive-trend test fails at two consecutive evaluations, which reduces sensitivity to a single noisy validation point.
We then restore the checkpoint with the highest primary validation metric observed in that stage and advance to the next stage.
The same rule is applied to validation Pass@1 in the final stage for early stopping.

\subsubsection{Stage-Aware Probability-Based Token Credit Redistribution}

Group-based RL assigns a trajectory-level advantage $A_i$ to all generated tokens in trajectory $\tau_i$.
The unweighted token contribution is
\[
g_{i,j}
=
A_i
\nabla_\theta
\log\pi_\theta
\left(
y_{i,j}\mid x_i,y_{i,<j}
\right).
\]
DHRCL redistributes this trajectory-level update across tokens:
\[
g'_{i,j}=\beta_{i,j}g_{i,j}.
\]
Let
\[
p_{i,j}
=
\pi_{\theta_{\mathrm{old}}}
\left(
y_{i,j}\mid x_i,y_{i,<j}
\right)
\]
be the detached rollout-policy probability of token $y_{i,j}$.
Token probability is interpreted as how established a decision is under the current policy, not as a token-correctness label.

During the Syntax Stage, we use confidence-oriented weighting,
\[
\beta_{i,j}^{\mathrm{conf}}
=
\frac{p_{i,j}}
{\frac{1}{|\tau_i|}\sum_{k=1}^{|\tau_i|}p_{i,k}}.
\]
For positive syntax advantages, this consolidates recurrent valid patterns; for negative advantages, it more strongly suppresses recurrent high-probability patterns associated with syntax failure.
The goal is to reduce the influence of incidental low-probability samples during early syntax-oriented optimization.

During the Execution Stage, we use
\[
\beta_{i,j}^{\mathrm{exec}}=1.
\]
Execution validity frequently depends on non-local interactions, and token probability alone does not reliably localize runtime responsibility.
Uniform weighting therefore avoids introducing an unsupported local bias.

During the Pass \& Structural Stage, we use uncertainty-oriented weighting,
\[
\beta_{i,j}^{\mathrm{unc}}
=
\frac{1-p_{i,j}}
{\frac{1}{|\tau_i|}\sum_{k=1}^{|\tau_i|}(1-p_{i,k})}.
\]
At this stage, the optimization bottleneck shifts toward decisions that remain less established under the current policy.
The weighting does not indiscriminately increase their probability: the sign of $A_i$ determines whether a low-probability decision receives stronger positive credit or stronger negative blame.

All three weighting rules have trajectory-level mean one:
\[
\frac{1}{|\tau_i|}\sum_{j=1}^{|\tau_i|}\beta_{i,j}=1.
\]
The mechanism therefore preserves the average contribution of a trajectory while changing how its gradient budget is allocated across generated tokens.
Overall, the design follows a consolidation-to-refinement principle: confidence-oriented consolidation in the syntax stage, uniform propagation for non-local execution feedback, and uncertainty-oriented refinement in the final stage.
Trajectory rewards are normalized within each prompt-level rollout group before advantage estimation.

\section{Experiments}
\label{sec:experiments}

We evaluate DHRCL under a controlled Qwen3-8B and KodCode setting and additionally examine scale robustness using Qwen3-4B, Qwen3-8B, and Qwen3-14B backbones.

\subsection{Experimental Setup}
\label{subsec:exp_setup}

All trainable methods are initialized from the corresponding backbone checkpoint and trained on KodCode~\citep{wang2026kodcode} under a matched RL protocol.
The detailed ablations use Qwen3-8B~\citep{yang2025qwen3}.
The cross-scale comparison uses Qwen3-4B, Qwen3-8B, and Qwen3-14B, which share the same dense Transformer family and tokenizer while differing in capacity.
A held-out KodCode validation split is used exclusively for trend-based stage progression, early stopping, and checkpoint selection.
Final evaluation uses HumanEval, HumanEval+, BigCodeBench-Full, BigCodeBench-Hard, LiveCodeBench V6, and CodeElo~\citep{codex,liu2023humanevalplus,bigcodebench,jain2024livecodebench,quan2025codeelo}.
Reference solutions are used only by the AST-based reward and are never included in rollout prompts.
All trainable configurations are repeated with three random seeds. Tables report mean $\pm$ standard deviation, and pairwise method differences are additionally assessed with problem-level paired bootstrap confidence intervals.

We compare DHRCL with GRPO, GRPO-PassRate, AceCoder, and VeRPO~\citep{dsmath,AceCoder,VeRPO}.
Each method retains its core reward formulation while using the same policy initialization, training corpus, rollout budget, optimization budget, decoding configuration, and execution sandbox at a given backbone scale.

Pass@1 is the primary functional metric.
Syntax Acc. and Exec Acc. describe intermediate program validity.
AST Sim., normalized execution time (NET), and normalized test memory usage (NTMU) are complementary diagnostics for accepted outputs.
Because NET and NTMU are computed over each method's own accepted subset, they are descriptive statistics rather than strictly paired efficiency comparisons.
For training-process analysis, we report Step Time, GPU Hours, Iterations to Target, Avg. DGR, and Reward Var.
Metric definitions and additional implementation details are provided in the Supplementary Document, Sections ``Program Validity and Accepted-Output Diagnostics'' and ``Training-Efficiency and Reward-Signal Metrics.''

\subsection{Comparison Experiments}
\label{subsec:comparison}

\begin{table*}[!t]
\centering
\scriptsize
\setlength{\tabcolsep}{1.8pt}
\renewcommand{\arraystretch}{0.92}
\caption{Main comparison under the unified Qwen3-8B and KodCode setting. Trainable results are mean $\pm$ standard deviation over three seeds.}
\label{tab:main_comparison}
\resizebox{\textwidth}{!}{
\begin{tabular}{l l c c c c c c c c c c c c c}
\toprule
\textbf{Method} & \textbf{Reward} & \textbf{Ext. RM}
& \textbf{HE} & \textbf{HE+} & \textbf{BCB-F} & \textbf{BCB-H} & \textbf{LCB} & \textbf{CodeElo} & \textbf{Avg.}
& \textbf{Syn. Acc.} & \textbf{Exec Acc.} & \textbf{AST Sim.} & \textbf{NET} & \textbf{NTMU} \\
\midrule
Qwen3-8B~\citep{yang2025qwen3} & -- & --
& 91.0 & 87.5 & 35.9 & 14.5 & 23.7 & 16.2 & 44.8
& 88.1 & 74.8 & 0.55 & 1.09$\times$ & 1.06$\times$ \\
GRPO~\citep{dsmath} & 0--1 & \xmark
& \ms{92.0}{0.3} & \ms{88.3}{0.3} & \ms{36.4}{0.4} & \ms{16.5}{0.3} & \ms{27.8}{0.4} & \ms{26.6}{0.6} & \ms{47.9}{0.2}
& \ms{89.1}{0.3} & \ms{77.7}{0.4} & \ms{0.57}{0.01} & \ms{1.00}{0.02}$\times$ & \ms{1.00}{0.02}$\times$ \\
GRPO-PassRate & Pass rate & \xmark
& \ms{92.4}{0.2} & \ms{88.5}{0.2} & \ms{37.0}{0.3} & \ms{16.9}{0.4} & \ms{28.6}{0.3} & \ms{27.0}{0.5} & \ms{48.4}{0.2}
& \ms{89.5}{0.2} & \ms{78.7}{0.3} & \ms{0.59}{0.01} & \ms{0.97}{0.02}$\times$ & \ms{0.99}{0.01}$\times$ \\
AceCoder~\citep{AceCoder} & RM dense & \cmark
& \ms{92.5}{0.3} & \ms{88.8}{0.3} & \ms{36.9}{0.4} & \ms{16.8}{0.5} & \ms{28.1}{0.5} & \ms{27.6}{0.7} & \ms{48.5}{0.3}
& \ms{89.8}{0.3} & \ms{79.1}{0.4} & \ms{0.60}{0.02} & \ms{0.95}{0.03}$\times$ & \ms{0.97}{0.02}$\times$ \\
VeRPO~\citep{VeRPO} & Verifiable dense & \xmark
& \ms{92.8}{0.2} & \ms{89.2}{0.2} & \ms{37.6}{0.3} & \ms{17.6}{0.3} & \ms{29.4}{0.4} & \ms{28.7}{0.6} & \ms{49.2}{0.2}
& \ms{90.5}{0.2} & \ms{80.3}{0.3} & \ms{0.63}{0.01} & \ms{0.92}{0.02}$\times$ & \ms{0.95}{0.02}$\times$ \\
\textbf{DHRCL} & Hierarchical dense & \xmark
& \textbf{\ms{93.1}{0.2}} & \textbf{\ms{89.8}{0.2}} & \textbf{\ms{38.3}{0.3}} & \textbf{\ms{18.6}{0.4}} & \textbf{\ms{30.5}{0.3}} & \textbf{\ms{31.4}{0.5}} & \textbf{\ms{50.3}{0.2}}
& \textbf{\ms{92.8}{0.2}} & \textbf{\ms{82.7}{0.3}} & \textbf{\ms{0.68}{0.01}} & \textbf{\ms{0.85}{0.02}$\times$} & \textbf{\ms{0.88}{0.02}$\times$} \\
\bottomrule
\end{tabular}}
\end{table*}

\begin{table*}[!t]
\centering
\scriptsize
\setlength{\tabcolsep}{2.4pt}
\renewcommand{\arraystretch}{0.92}
\caption{Training-efficiency and reward-signal statistics on Qwen3-8B. Values are mean $\pm$ standard deviation over three seeds.}
\label{tab:main_training_logs}
\resizebox{\textwidth}{!}{
\begin{tabular}{l c c c c c c}
\toprule
\textbf{Method}
& \textbf{Ext. RM}
& \textbf{Step Time $\downarrow$}
& \textbf{GPU Hours $\downarrow$}
& \textbf{Iter. to Target $\downarrow$}
& \textbf{Avg. DGR $\downarrow$}
& \textbf{Reward Var.} \\
\midrule

GRPO~\citep{dsmath}
& \xmark
& \textbf{\ms{1.00}{0.02}$\times$}
& \ms{1.00}{0.04}$\times$
& \ms{1.00}{0.05}$\times$
& \ms{0.66}{0.02}
& \ms{0.035}{0.003} \\

GRPO-PassRate
& \xmark
& \ms{1.02}{0.02}$\times$
& \ms{0.81}{0.04}$\times$
& \ms{0.79}{0.05}$\times$
& \ms{0.43}{0.02}
& \ms{0.085}{0.006} \\

AceCoder~\citep{AceCoder}
& \cmark
& \ms{1.61}{0.05}$\times$
& \ms{1.34}{0.08}$\times$
& \ms{0.84}{0.07}$\times$
& \textbf{\ms{0.14}{0.02}}
& \ms{0.158}{0.011} \\

VeRPO~\citep{VeRPO}
& \xmark
& \ms{1.05}{0.02}$\times$
& \ms{0.69}{0.04}$\times$
& \ms{0.66}{0.04}$\times$
& \ms{0.25}{0.02}
& \ms{0.103}{0.007} \\

\textbf{DHRCL}
& \xmark
& \ms{1.11}{0.03}$\times$
& \textbf{\ms{0.55}{0.03}$\times$}
& \textbf{\ms{0.49}{0.04}$\times$}
& \underline{\ms{0.20}{0.01}}
& \ms{0.113}{0.008} \\

\bottomrule
\end{tabular}}
\end{table*}

\begin{table*}[!t]
\centering
\scriptsize
\setlength{\tabcolsep}{2.1pt}
\renewcommand{\arraystretch}{0.92}
\caption{Cross-scale Pass@1 comparison across Qwen3-4B, Qwen3-8B, and Qwen3-14B. Trainable results are mean $\pm$ standard deviation over three seeds.}
\label{tab:cross_scale}
\resizebox{\textwidth}{!}{
\begin{tabular}{l l c c c c c c c}
\toprule
\textbf{Backbone} & \textbf{Method}
& \textbf{HE} & \textbf{HE+} & \textbf{BCB-F} & \textbf{BCB-H}
& \textbf{LCB} & \textbf{CodeElo} & \textbf{Avg.}\\
\midrule
\multirow{6}{*}{Qwen3-4B}
& Base & 86.2 & 81.9 & 28.5 & 10.7 & 17.6 & 12.4 & 39.6\\
& GRPO & \ms{88.5}{0.4} & \ms{84.4}{0.4} & \ms{31.5}{0.5} & \ms{12.8}{0.4} & \ms{21.6}{0.5} & \ms{19.8}{0.8} & \ms{43.1}{0.3}\\
& GRPO-PassRate & \ms{88.9}{0.3} & \ms{84.8}{0.3} & \ms{31.9}{0.4} & \ms{13.1}{0.5} & \ms{22.4}{0.4} & \ms{20.7}{0.7} & \ms{43.6}{0.3}\\
& AceCoder & \ms{89.2}{0.4} & \ms{84.9}{0.4} & \ms{32.3}{0.5} & \ms{12.9}{0.6} & \ms{22.0}{0.6} & \ms{21.2}{0.8} & \ms{43.8}{0.4}\\
& VeRPO & \ms{89.6}{0.3} & \ms{85.4}{0.3} & \ms{32.9}{0.4} & \ms{13.8}{0.4} & \ms{23.2}{0.5} & \ms{22.0}{0.7} & \ms{44.5}{0.3}\\
& \textbf{DHRCL} & \textbf{\ms{90.3}{0.3}} & \textbf{\ms{86.4}{0.3}} & \textbf{\ms{34.0}{0.4}} & \textbf{\ms{14.7}{0.5}} & \textbf{\ms{24.8}{0.4}} & \textbf{\ms{24.3}{0.6}} & \textbf{\ms{45.8}{0.3}}\\
\midrule
\multirow{6}{*}{Qwen3-8B}
& Base & 91.0 & 87.5 & 35.9 & 14.5 & 23.7 & 16.2 & 44.8\\
& GRPO & \ms{92.0}{0.3} & \ms{88.3}{0.3} & \ms{36.4}{0.4} & \ms{16.5}{0.3} & \ms{27.8}{0.4} & \ms{26.6}{0.6} & \ms{47.9}{0.2}\\
& GRPO-PassRate & \ms{92.4}{0.2} & \ms{88.5}{0.2} & \ms{37.0}{0.3} & \ms{16.9}{0.4} & \ms{28.6}{0.3} & \ms{27.0}{0.5} & \ms{48.4}{0.2}\\
& AceCoder & \ms{92.5}{0.3} & \ms{88.8}{0.3} & \ms{36.9}{0.4} & \ms{16.8}{0.5} & \ms{28.1}{0.5} & \ms{27.6}{0.7} & \ms{48.5}{0.3}\\
& VeRPO & \ms{92.8}{0.2} & \ms{89.2}{0.2} & \ms{37.6}{0.3} & \ms{17.6}{0.3} & \ms{29.4}{0.4} & \ms{28.7}{0.6} & \ms{49.2}{0.2}\\
& \textbf{DHRCL} & \textbf{\ms{93.1}{0.2}} & \textbf{\ms{89.8}{0.2}} & \textbf{\ms{38.3}{0.3}} & \textbf{\ms{18.6}{0.4}} & \textbf{\ms{30.5}{0.3}} & \textbf{\ms{31.4}{0.5}} & \textbf{\ms{50.3}{0.2}}\\
\midrule
\multirow{6}{*}{Qwen3-14B}
& Base & 94.5 & 91.6 & 43.7 & 22.8 & 34.1 & 28.9 & 52.6\\
& GRPO & \ms{95.2}{0.2} & \ms{92.3}{0.2} & \ms{45.0}{0.3} & \ms{24.7}{0.4} & \ms{36.7}{0.4} & \ms{34.5}{0.6} & \ms{54.7}{0.2}\\
& GRPO-PassRate & \ms{95.5}{0.2} & \ms{92.5}{0.2} & \ms{45.5}{0.3} & \ms{25.1}{0.4} & \ms{37.1}{0.4} & \ms{35.0}{0.6} & \ms{55.1}{0.2}\\
& AceCoder & \ms{95.6}{0.2} & \ms{92.7}{0.2} & \ms{45.7}{0.4} & \ms{25.0}{0.5} & \ms{37.4}{0.5} & \ms{35.3}{0.7} & \ms{55.3}{0.3}\\
& VeRPO & \textbf{\ms{95.9}{0.2}} & \ms{92.9}{0.2} & \ms{46.2}{0.3} & \ms{25.8}{0.4} & \ms{38.0}{0.4} & \ms{36.5}{0.6} & \ms{55.9}{0.2}\\
& \textbf{DHRCL} & \ms{95.8}{0.2} & \textbf{\ms{93.2}{0.2}} & \textbf{\ms{46.8}{0.3}} & \textbf{\ms{26.7}{0.4}} & \textbf{\ms{38.9}{0.4}} & \textbf{\ms{38.4}{0.5}} & \textbf{\ms{56.6}{0.2}}\\
\bottomrule
\end{tabular}}
\end{table*}

Table~\ref{tab:main_comparison} reports the controlled Qwen3-8B comparison.
DHRCL achieves an average Pass@1 of $50.3\pm0.2$, outperforming VeRPO, the strongest baseline, by 1.1 points. A problem-level paired bootstrap comparison gives a 95\% confidence interval of $[0.6,1.5]$ for this average improvement.
The gain is accompanied by higher syntax and execution validity, with Syntax Acc. and Exec Acc. reaching $92.8\pm0.2$ and $82.7\pm0.3$, respectively.
DHRCL also obtains the highest mean AST similarity, while its accepted outputs exhibit lower average NET and NTMU under the adopted evaluation protocol.

Because all baselines are retrained under the unified Qwen3-8B and
KodCode protocol, the comparison reflects their adapted performance
under matched data, model, and training budgets rather than their
originally reported absolute results.
Table~\ref{tab:cross_scale} shows that DHRCL retains the highest average Pass@1 across Qwen3-4B, Qwen3-8B, and Qwen3-14B. Its average improvement over VeRPO is 1.3, 1.1, and 0.7 points, with paired-bootstrap 95\% confidence intervals of $[0.8,1.8]$, $[0.6,1.5]$, and $[0.3,1.2]$, respectively. The observed margin narrows with model capacity while remaining positive under the reported paired-bootstrap analysis. Results are not uniformly dominant on every benchmark: for Qwen3-14B, VeRPO is marginally higher on HumanEval, while DHRCL retains larger gains on BigCodeBench-Hard, LiveCodeBench, and CodeElo.

\subsection{Ablation Studies}
\label{subsec:ablations}

This subsection isolates the three main components of DHRCL: reward composition, curriculum scheduling, and probability-aware token credit redistribution.
Table~\ref{tab:ablation_summary} summarizes controlled ablations for reward composition, curriculum scheduling, and probability-aware token credit redistribution.

\begin{table*}[!t]
\centering
\scriptsize
\setlength{\tabcolsep}{3.2pt}
\renewcommand{\arraystretch}{0.92}
\caption{Controlled ablations on Qwen3-8B. Avg. Pass@1 is reported as mean $\pm$ standard deviation over three seeds.}
\label{tab:ablation_summary}
\resizebox{\textwidth}{!}{
\begin{tabular}{l l c c c c c c c c}
\toprule
\textbf{Group} & \textbf{Variant}
& \textbf{Avg.} & \textbf{Syn. Acc.} & \textbf{Exec Acc.} & \textbf{AST Sim.}
& \textbf{NET} & \textbf{NTMU}
& \textbf{Iter. to Target} & \textbf{Avg. DGR $\downarrow$} \\
\midrule
\multirow{5}{*}{Reward}
& Pass only & \ms{48.4}{0.3} & 89.3 & 78.5 & 0.58 & 0.98$\times$ & 0.99$\times$ & 0.78$\times$ & 0.43\\
& Pass + Syntax & \ms{49.0}{0.3} & 91.7 & 79.3 & 0.60 & 0.95$\times$ & 0.97$\times$ & 0.67$\times$ & 0.32\\
& Pass + Exec & \ms{48.8}{0.3} & 89.9 & 81.7 & 0.60 & 0.94$\times$ & 0.96$\times$ & 0.69$\times$ & 0.30\\
& Pass + Syntax + Exec & \ms{49.7}{0.2} & 92.4 & 82.1 & 0.62 & 0.91$\times$ & 0.94$\times$ & 0.59$\times$ & 0.26\\
& Full Reward & \textbf{\ms{50.3}{0.2}} & \textbf{92.8} & \textbf{82.7} & \textbf{0.68} & \textbf{0.85$\times$} & \textbf{0.88$\times$} & \textbf{0.49$\times$} & \textbf{0.20}\\
\midrule
\multirow{3}{*}{Curriculum}
& Fixed Sum & \ms{49.1}{0.3} & 91.5 & 80.8 & 0.64 & 0.91$\times$ & 0.94$\times$ & 0.70$\times$ & 0.28\\
& Fixed-stage & \ms{49.6}{0.3} & 92.1 & 81.9 & 0.66 & 0.88$\times$ & 0.91$\times$ & 0.58$\times$ & 0.23\\
& Trend-based DHRCL & \textbf{\ms{50.3}{0.2}} & \textbf{92.8} & \textbf{82.7} & \textbf{0.68} & \textbf{0.85$\times$} & \textbf{0.88$\times$} & \textbf{0.49$\times$} & \textbf{0.20}\\
\midrule
\multirow{5}{*}{Prob.-aware}
& No Weighting & \ms{49.4}{0.3} & 92.0 & 81.8 & 0.65 & 0.90$\times$ & 0.92$\times$ & 0.64$\times$ & 0.25\\
& Early-only & \ms{49.8}{0.2} & 92.7 & 82.0 & 0.66 & 0.88$\times$ & 0.91$\times$ & 0.56$\times$ & 0.22\\
& Late-only & \ms{49.9}{0.3} & 92.2 & 82.5 & 0.67 & 0.87$\times$ & 0.90$\times$ & 0.54$\times$ & 0.21\\
& Reversed & \ms{49.2}{0.3} & 91.6 & 81.4 & 0.64 & 0.92$\times$ & 0.94$\times$ & 0.68$\times$ & 0.27\\
& Full DHRCL & \textbf{\ms{50.3}{0.2}} & \textbf{92.8} & \textbf{82.7} & \textbf{0.68} & \textbf{0.85$\times$} & \textbf{0.88$\times$} & \textbf{0.49$\times$} & \textbf{0.20}\\
\bottomrule
\end{tabular}}
\end{table*}

\paragraph{Reward hierarchy.}
For the reward-component ablation, the curriculum rule,
probability-aware token weighting, optimizer, rollout budget, and
training budget are held fixed; only the included reward components
change. The comparison between \textit{Pass + Syntax + Exec} and
\textit{Full Reward} isolates the AST-based structural reward.
Adding the AST reward increases average Pass@1 from 49.7 to 50.3 and
AST similarity from 0.62 to 0.68. We regard the structural-alignment
gain as its primary effect, while the smaller Pass@1 improvement
suggests a secondary functional benefit. NET and NTMU are reported as
descriptive statistics of accepted outputs because the solved-problem
subsets differ across variants.

\paragraph{Curriculum scheduling.}
All curriculum variants use the same full reward, probability-aware weighting, optimizer, rollout budget, and training budget.
They differ only in reward scheduling and stage progression.
\textit{Fixed Sum} jointly optimizes all components from the beginning, \textit{Fixed-stage} changes stages after predefined training iterations, and DHRCL uses the recent validation trend.
This comparison isolates whether automatically adapting stage duration is preferable to static aggregation or a fixed schedule.

\paragraph{Probability-aware token credit redistribution.}
All token-weighting variants use the same hierarchical reward and trend-based curriculum.
They differ only in how the trajectory-level gradient budget is distributed among generated tokens.
No Weighting uses uniform propagation in all stages; Early-only and
Late-only apply one side of the proposed ordering; and Reversed applies
uncertainty weighting early and confidence weighting late. The full
method evaluates the proposed consolidation-to-uniform-to-refinement
ordering. The comparison therefore isolates whether stage-dependent
token weighting improves the allocation of the trajectory-level
gradient budget.

\subsection{Further Analysis}
\label{subsec:further_analysis}

We use Qwen3-8B for detailed diagnostics because it is the primary ablation setting; cross-scale effectiveness is established in Table~\ref{tab:cross_scale}.

\begin{figure}[!t]
\centering
\includegraphics[width=0.95\linewidth]{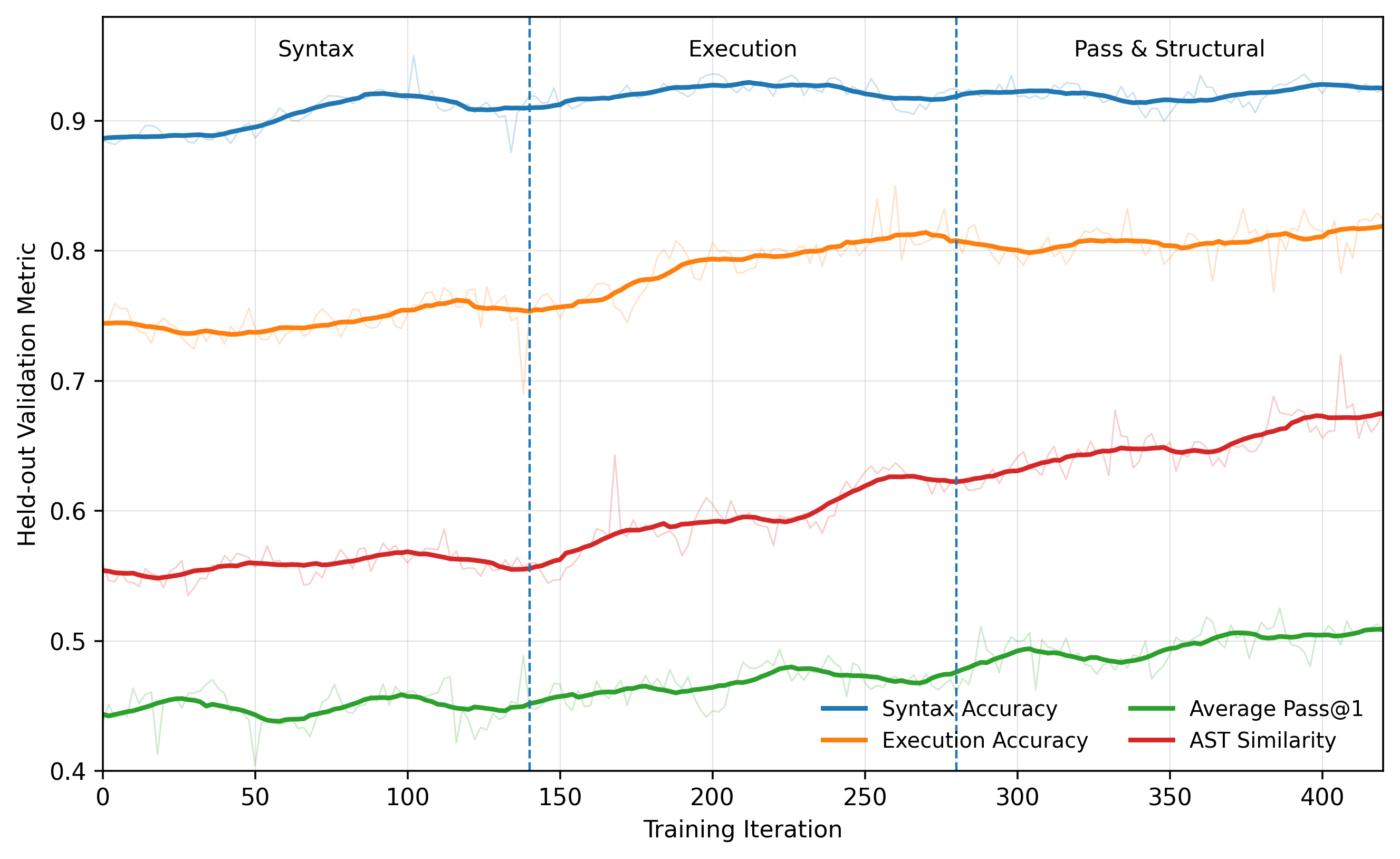}
\caption{Training dynamics of DHRCL on Qwen3-8B.}
\label{fig:training_dynamics}
\end{figure}

\paragraph{Training dynamics.}
Figure~\ref{fig:training_dynamics} shows the evolution of held-out
validation metrics throughout DHRCL training on Qwen3-8B.
Syntax Accuracy improves most rapidly in the first stage, Execution
Accuracy rises more clearly after the first transition, and average
Pass@1 and AST Similarity obtain their largest gains after the second
transition. Raw measurements fluctuate because of rollout and
optimization stochasticity, whereas the moving averages reveal the
longer-term stage-wise progression. Across three seeds, the median
transition iterations are 140/280 for Qwen3-8B, 180/340 for Qwen3-4B,
and 120/260 for Qwen3-14B, indicating that the selected stage duration
varies with model scale.

\begin{figure}[!t]
\centering
\includegraphics[width=0.95\linewidth]{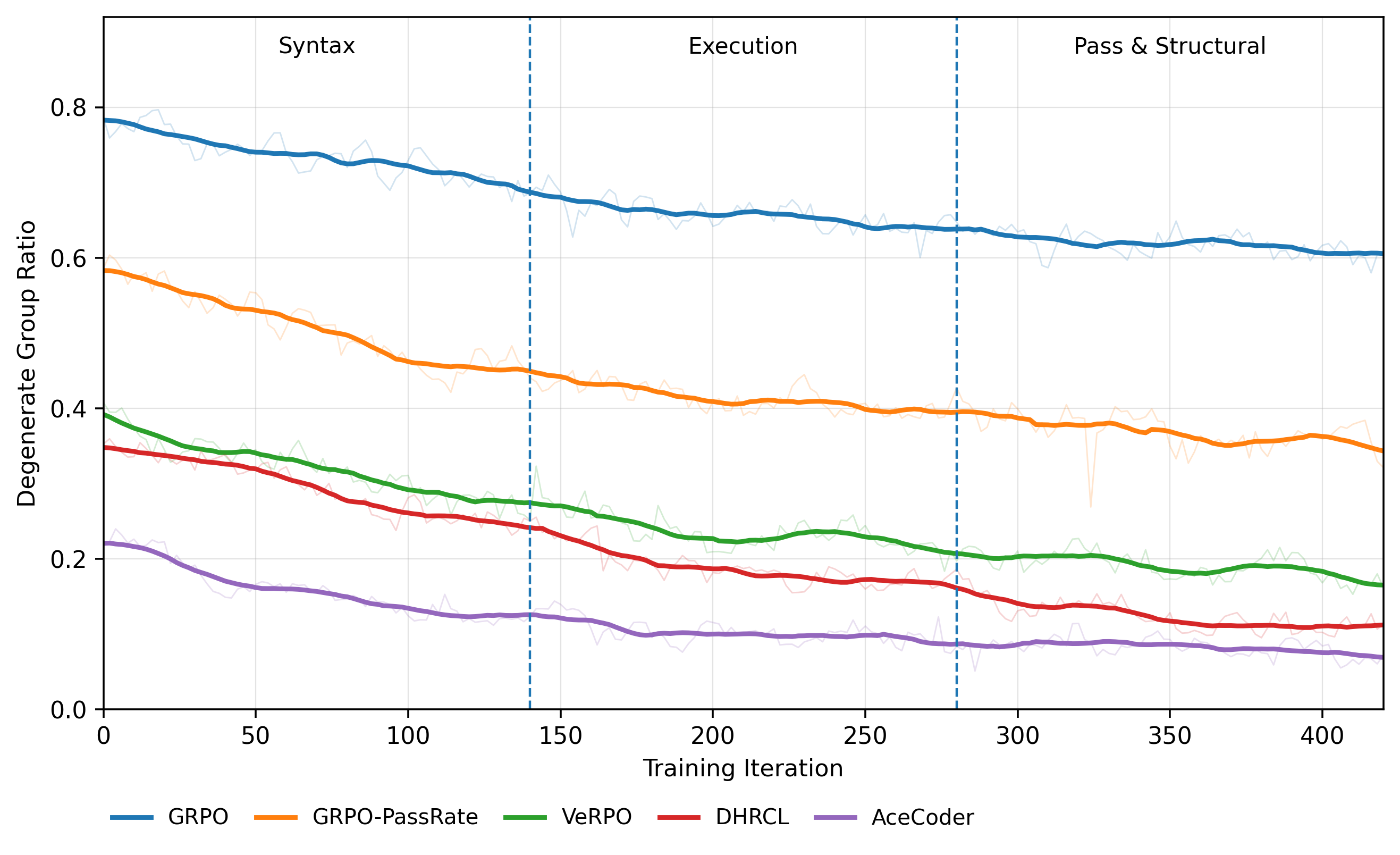}
\caption{Degenerate group ratio on Qwen3-8B.}
\label{fig:dgr_curve}
\end{figure}

\paragraph{Reward-signal discriminability.}
DGR measures how often a rollout group receives nearly identical rewards and therefore little relative-advantage information.
As shown in Figure~\ref{fig:dgr_curve}, binary GRPO has the highest DGR, partial-test and verifiable rewards reduce it, and AceCoder is lowest because its learned reward model produces continuously varying scores.
DHRCL achieves the second-lowest DGR without an external reward model,
indicating that its hierarchical feedback distinguishes more candidates
within each rollout group than the binary, pass-rate, and verifiable
dense-reward alternatives considered here.

Additional error-type breakdowns are provided in the Supplementary Document, Section ``Error-Type Analysis.'' They show that DHRCL reduces invalid programs while executable-but-incorrect solutions remain the dominant residual failure mode.

\section{Conclusion}
We introduced DHRCL, which coordinates syntax, execution, test, and AST-based structural feedback through a trend-adaptive curriculum and stage-aware token credit redistribution.
Controlled Qwen3-8B experiments and a Qwen3-4B/8B/14B scale study show consistent gains in functional correctness, intermediate validity, structural alignment, reward discriminability, and convergence efficiency.

\section*{Limitations}
AST similarity is an auxiliary reference-conditioned signal rather than semantic equivalence; token probability does not identify semantic correctness; the three-stage capability order remains manually specified; and the scale study is limited to the Qwen3 family.
NET and NTMU are descriptive statistics over each method's accepted subset.
Broader cross-family and cross-corpus validation remains future work.

\clearpage
\appendix
\section*{Supplementary Material}
\addcontentsline{toc}{section}{Supplementary Material}

\captionsetup[table]{font=scriptsize,skip=2pt}
\setlength{\abovecaptionskip}{3pt}
\setlength{\belowcaptionskip}{2pt}
\setlength{\dbltextfloatsep}{8pt plus 2pt minus 2pt}
\setlength{\dblfloatsep}{8pt plus 2pt minus 2pt}
\raggedbottom

\section*{Overview}
This supplementary document provides additional implementation details,
evaluation protocols, metric definitions, and extended ablation results.

\section{Baseline Adaptation Protocol}

All trainable baselines are retrained under a matched protocol rather
than evaluated using released policy checkpoints. At each backbone
scale, GRPO, GRPO-PassRate, VeRPO, AceCoder, and DHRCL use the same
policy initialization, KodCode training split, rollout budget,
optimization budget, decoding configuration, execution sandbox, and
checkpoint-selection protocol. The reported results therefore
characterize these unified adaptations rather than reproducing the
original papers' absolute performance.

Each baseline retains its defining reward formulation. GRPO uses binary
complete-test outcomes, GRPO-PassRate uses the fraction of passed test
cases, VeRPO uses its difficulty-aware verifiable reward and policy
objective, and AceCoder uses the released AceCodeRM as an external
reward model. Baseline-specific hyperparameters and reward
normalization are selected under the same held-out validation protocol
and comparable tuning budgets.

\section{Training Configuration}
\label{app:training_config}

Experiments are conducted on 8 NVIDIA A100 80GB GPUs for the Qwen3-8B setting.
The Qwen3-4B experiments use 4 NVIDIA A100 80GB GPUs, while the Qwen3-14B experiments use 16 NVIDIA A100 80GB GPUs.
Each rollout batch contains 32 programming problems with 8 candidate solutions per problem.
Training-time generation uses temperature $T=1.0$, top-$p=1.0$, top-$k=-1$, and a maximum response length of 16384 tokens.
Evaluation uses the protocol described below under \textit{Benchmark Evaluation Protocol}.

All methods use the same maximum training-iteration budget at a given backbone scale. Each trainable configuration is run with three random seeds.
The main performance results report mean and standard
deviation, while supplementary diagnostic tables report seed
means unless otherwise stated.
All policies are optimized with full-parameter AdamW using a peak learning rate of $1\times10^{-6}$, $\beta_1=0.9$, $\beta_2=0.95$, and weight decay $0.1$. We use a 3\% linear warmup followed by cosine decay, gradient clipping at a global norm of 1.0, one policy-update epoch per rollout batch, bfloat16 mixed precision, and FSDP full sharding. The same optimizer and stabilization settings are used for all trainable methods at a given backbone scale.

For GRPO-style updates, we use Clip-Higher with $\epsilon_{\mathrm{low}}=0.2$ and $\epsilon_{\mathrm{high}}=0.28$ and omit an explicit KL penalty.
For VeRPO, the adapted implementation uses $\alpha=2.0$ and $\beta=1.0$; the trajectory-length decay is inactive in the single-turn setting.

For DHRCL, validation is performed every two training iterations on a fixed held-out validation set using a fixed decoding protocol and fixed random seeds. After at least 12 validation observations have been collected in the current stage, a one-sided linear-trend test at significance level $0.05$ is applied to the most recent eight observations. A stage transition is triggered only when the test fails to support a positive trend at two consecutive evaluations. The best checkpoint of the current stage is then restored before training advances. The monitored metric is syntax reward in the Syntax Stage, execution reward in the Execution Stage, and validation Pass@1 in the final stage. All curriculum, reward, and token-weighting ablations use this same rule.

\section{Dataset Splits}
\label{app:dataset_split}

We use KodCode as the RL training corpus.
The KodCode training split is used for rollout generation, reward computation, and policy optimization.
From the training corpus, we reserve a held-out validation split for curriculum-stage switching, early stopping, reward-curve logging, and training-stability analysis.
The validation split is never used for policy-gradient updates and is not included in final benchmark reporting.

Final evaluation is conducted only on external benchmark suites, including HumanEval, HumanEval+, BigCodeBench-Full, BigCodeBench-Hard, LiveCodeBench V6, and CodeElo.
These benchmarks serve as the test sets for reporting final Pass@1, intermediate validity, and accepted-output diagnostic metrics.
This separation ensures that curriculum decisions are made on held-out in-domain validation problems, while final results are measured on external evaluation benchmarks.

Reference solutions in KodCode are used only for computing reward components that require structural comparison, such as AST similarity.
They are not included in model inputs and are not exposed to the policy during rollout generation.
During rollout generation, the policy only receives the programming problem prompt and generates candidate solutions under the specified decoding configuration.

\section{Benchmark Evaluation Protocol}

All methods are evaluated under the same benchmark protocol, decoding configuration, and execution environment.
During benchmark evaluation, we sample $n=8$ candidate solutions for each problem and report Pass@1.
Evaluation-time generation uses temperature $T=0.6$, top-$p=0.95$, and top-$k=20$.
The maximum response length is set to 16384 tokens for both training and evaluation, following the VeRPO-aligned protocol and reducing truncation effects on long-form code-generation tasks.
All generated programs are executed in the same sandbox with fixed timeout, memory constraints, and unit-test protocol.

For each benchmark, the same prompt format, decoding parameters, execution sandbox, and pass/fail criteria are used across all methods.
For method comparisons, we compute problem-level paired bootstrap confidence intervals with 10,000 resamples, preserving the pairing of predictions on each benchmark problem.
This ensures that differences in final results reflect differences in training objectives rather than differences in evaluation configuration.
NET and NTMU are computed only on accepted programs. Because each method may solve a different subset of problems, they are interpreted as descriptive accepted-output diagnostics rather than strictly paired efficiency comparisons.

\section{Program Validity and Accepted-Output Diagnostics}
\label{app:quality_metrics}

Pass@1 is the primary functional metric.
Syntax Acc. is a binary program-level metric obtained with a strict language parser, whereas the training-time syntax reward provides partial credit over local units using an error-tolerant parser.
Exec Acc. is the fraction of generated programs that finish without runtime exceptions, timeout, or memory-limit failure.

AST Sim. is computed between a generated program and a verified reference after applying the exact normalization and similarity procedure specified below under \textit{Implementation Details of the AST Structural Reward}.
The metric describes reference-conditioned structural alignment and is not interpreted as semantic equivalence.
A generated program that cannot be strictly parsed receives zero AST reward during training.
\begin{table*}[!t]
\centering
\scriptsize
\setlength{\tabcolsep}{3.2pt}
\renewcommand{\arraystretch}{0.90}

\refstepcounter{table}
\label{tab:error_distribution}
\textbf{Table~\thetable: Error-type distribution under the unified Qwen3-8B evaluation protocol.
All categories are mutually exclusive, and each row sums to 100\%.}\\[2pt]
\begin{tabular}{l c c c c c}
\toprule
\textbf{Method}
& \textbf{SyntaxError (\%) $\downarrow$}
& \textbf{RuntimeError (\%) $\downarrow$}
& \textbf{Timeout (\%) $\downarrow$}
& \textbf{Wrong Answer (\%) $\downarrow$}
& \textbf{Accepted (\%) $\uparrow$} \\
\midrule
Qwen3-8B~\citep{yang2025qwen3} & 8.0 & 9.7 & 4.8 & 32.7 & 44.8 \\
GRPO~\citep{dsmath} & 7.3 & 8.7 & 4.1 & 32.0 & 47.9 \\
GRPO-PassRate & 6.8 & 8.1 & 3.8 & 32.9 & 48.4 \\
AceCoder~\citep{AceCoder} & 7.1 & 8.3 & 4.0 & 32.1 & 48.5 \\
VeRPO~\citep{VeRPO} & 6.2 & 7.4 & 3.5 & 33.7 & 49.2 \\
\textbf{DHRCL} & \textbf{4.8} & \textbf{5.9} & \textbf{2.9} & 36.1 & \textbf{50.3} \\
\bottomrule
\end{tabular}

\vspace{7pt}

\refstepcounter{table}
\label{tab:reward_ablation_full}
\textbf{Table~\thetable: Reward-component ablation.}\\[2pt]
\resizebox{\textwidth}{!}{
\begin{tabular}{l c c c c c c c c c c c c c c c c}
\toprule
\textbf{Variant} & \textbf{Syn.} & \textbf{Exec} & \textbf{Pass} & \textbf{AST}
& \textbf{HE} & \textbf{HE+} & \textbf{BCB-F} & \textbf{BCB-H} & \textbf{LCB} & \textbf{CodeElo} & \textbf{Avg.}
& \textbf{Syn. Acc.} & \textbf{Exec Acc.} & \textbf{AST Sim.} & \textbf{NET} & \textbf{NTMU}\\
\midrule
Pass only & \xmark & \xmark & \cmark & \xmark & 92.3 & 88.5 & 36.9 & 16.8 & 28.4 & 27.5 & 48.4 & 89.3 & 78.5 & 0.58 & 0.98$\times$ & 0.99$\times$\\
Pass + Syntax & \cmark & \xmark & \cmark & \xmark & 92.7 & 89.0 & 37.3 & 17.3 & 29.0 & 28.7 & 49.0 & 91.7 & 79.3 & 0.60 & 0.95$\times$ & 0.97$\times$\\
Pass + Exec & \xmark & \cmark & \cmark & \xmark & 92.5 & 88.8 & 37.1 & 17.2 & 29.0 & 27.9 & 48.8 & 89.9 & 81.7 & 0.60 & 0.94$\times$ & 0.96$\times$\\
Pass + Syntax + Exec & \cmark & \cmark & \cmark & \xmark & 92.8 & 89.3 & 37.7 & 18.0 & 29.8 & 30.6 & 49.7 & 92.4 & 82.1 & 0.62 & 0.91$\times$ & 0.94$\times$\\
Full Reward & \cmark & \cmark & \cmark & \cmark & \textbf{93.1} & \textbf{89.8} & \textbf{38.3} & \textbf{18.6} & \textbf{30.5} & \textbf{31.4} & \textbf{50.3} & \textbf{92.8} & \textbf{82.7} & \textbf{0.68} & \textbf{0.85$\times$} & \textbf{0.88$\times$}\\
\bottomrule
\end{tabular}}

\vspace{7pt}

\refstepcounter{table}
\label{tab:curriculum_prob_ablation_full}
\textbf{Table~\thetable: Curriculum and token-credit ablations.}\\[2pt]
\begin{tabular}{l l c c c c c}
\toprule
\textbf{Group} & \textbf{Variant} & \textbf{Avg. Pass@1} & \textbf{Iterations to Target} & \textbf{Avg. DGR} & \textbf{GPU Hours} & \textbf{Stage Transitions}\\
\midrule
Curriculum & Fixed Sum & \ms{49.1}{0.3} & 0.70$\times$ & 0.28 & 0.78$\times$ & --\\
Curriculum & Fixed-stage & \ms{49.6}{0.3} & 0.58$\times$ & 0.23 & 0.65$\times$ & 160/300\\
Curriculum & Trend-based DHRCL & \textbf{\ms{50.3}{0.2}} & \textbf{0.49$\times$} & \textbf{0.20} & \textbf{0.55$\times$} & 140/280\\
\midrule
Token & No Weighting & \ms{49.4}{0.3} & 0.64$\times$ & 0.25 & 0.72$\times$ & 160/300\\
Token & Early-only & \ms{49.8}{0.2} & 0.56$\times$ & 0.22 & 0.63$\times$ & 140/300\\
Token & Late-only & \ms{49.9}{0.3} & 0.54$\times$ & 0.21 & 0.61$\times$ & 160/280\\
Token & Reversed & \ms{49.2}{0.3} & 0.68$\times$ & 0.27 & 0.76$\times$ & 180/320\\
Token & Full DHRCL & \textbf{\ms{50.3}{0.2}} & \textbf{0.49$\times$} & \textbf{0.20} & \textbf{0.55$\times$} & 140/280\\
\bottomrule
\end{tabular}
\end{table*}

NET is the execution time of an accepted generated program normalized by the corresponding aggregate GRPO value.
NTMU is the peak test memory usage of an accepted generated program normalized in the same way.
Both metrics are computed over each method's own accepted outputs.
Because the solved-problem subsets can differ, NET and NTMU are descriptive computational characteristics under the adopted evaluation protocol, not strictly paired evidence that one method is universally more efficient.

We keep these evidence levels separate:
Pass@1 measures functional correctness;
Syntax Acc. and Exec Acc. diagnose intermediate validity;
AST Sim. measures structural alignment;
and NET/NTMU describe the observed computation of accepted outputs.

\section{Training-Efficiency and Reward-Signal Metrics}
\label{app:training_log_metrics}

We report training-log metrics to compare computational cost, convergence efficiency, reward informativeness, and reward dispersion across RL methods.

Step Time is the average wall-clock time per training iteration, including rollout generation, execution, reward computation, log-probability recomputation, and policy update.
GPU Hours measures the GPU-side cost required to reach the target validation performance.
It is computed from the average wall-clock time per training iteration, the number of iterations required to reach the target, and the number of GPUs used during training.
All values are normalized by GRPO.
For methods that do not reach the target within the fixed training budget, we report the GPU Hours consumed under the full budget and mark Iterations to Target as N.R.

Iterations to Target measures how many training iterations are required to reach a fixed validation-performance threshold.
We set this threshold to the final validation Pass@1 achieved by GRPO under the same training budget.
GRPO therefore serves as the reference method with normalized Iterations to Target equal to $1.00\times$.
For methods that do not reach this threshold within the training budget, we report N.R.
A lower value indicates faster convergence to the same validation-performance level.

Avg. DGR denotes the average degenerate group ratio over logged training iterations.
For each prompt, the policy samples a rollout group $G=\{\tau_i\}_{i=1}^{N}$ with scalar rewards $\{R(\tau_i)\}_{i=1}^{N}$.
A rollout group is considered degenerate if
\[
\mathrm{Std}\left(\{R(\tau_i)\}_{i=1}^{N}\right)<10^{-2}.
\]
The degenerate group ratio at a training iteration is the proportion of degenerate groups among all prompt-level rollout groups in that iteration.
Avg. DGR is the average of this ratio over logged training iterations.
A lower Avg. DGR indicates that a larger proportion of rollout groups provide non-trivial relative-advantage signals for policy optimization.

Reward Var. denotes the average within-group reward variance over logged training iterations.
It is used to characterize reward dispersion rather than as a metric to be minimized.
Extremely low reward variance often indicates uninformative rewards, while excessively high variance may make optimization unstable.

In the reward-signal analysis, DGR is used to compare how often different reward designs fail to distinguish sampled candidates within the same prompt-level rollout group.
Binary pass/fail rewards tend to produce high DGR when most sampled programs fail all tests.
Pass-rate and difficulty-aware rewards reduce this issue by assigning different scores to partially correct programs.
AceCoder achieves the lowest DGR through its external learned reward model. DHRCL achieves the second-lowest DGR without an external reward model because programs may differ in syntax validity, execution status, unit-test pass rate, and structural similarity.
This makes the rollout group more likely to provide useful relative-advantage signals during group-based policy optimization.

\section{Implementation Details of the AST Structural Reward}

The AST reward is implemented for Python programs using Python 3.11's built-in \texttt{ast} parser. Before comparison, Markdown fences, comments, docstrings, formatting information, and source-location attributes are removed. User-defined identifiers are consistently renamed in order of first occurrence, while operators, control-flow node types, library API names, and the constants $-1$, $0$, and $1$ are preserved. Other string and numeric literals are mapped to type-level placeholders.

For normalized trees $A$ and $B$, structural similarity is computed from unit-cost tree-edit distance:
\[
\mathrm{Sim}_{\mathrm{AST}}(A,B)
=
\max\left(0,1-\frac{\mathrm{TED}(A,B)}{|A|+|B|}\right),
\]
where $|A|$ and $|B|$ denote the numbers of AST nodes. Each problem uses its verified reference solution; when multiple verified references are available, the maximum similarity is used.

For unparsable programs, the training reward is zero.
For parsable programs, the returned similarity is bounded in $[0,1]$.
This signal is used only as reference-conditioned structural supervision; unit-test feedback remains the direct functional objective.

\section{Implementation Details of Syntax Reward}
\label{app:syntax_reward_impl}

During RL training, generated programs often contain local syntax errors, so a strict AST parser may provide no partial feedback.
We therefore use Tree-sitter as an error-tolerant parser to obtain partial syntax trees after removing Markdown code fences and normalizing indentation when possible.

We extract statement- and block-level syntactic units from the parse tree, including function definitions, class definitions, control-flow statements, assignments, expressions, and return statements.
Lexical leaves, comments, and overly short fragments are excluded from reward computation.
A unit is marked as syntactically valid if its subtree does not contain parser-reported error nodes.
The final syntax reward is the length-weighted fraction of valid units:
\[
r_{\mathrm{syntax}}(y)=
\frac{\sum_{i=1}^{m} w_i v_i}
{\sum_{i=1}^{m} w_i}.
\]
We use token length as $w_i$ so that larger program regions contribute proportionally more to the reward.
If the parser cannot extract any statement-level unit, the syntax reward is set to zero.

This implementation makes the training reward robust to early-stage generations that are only partially well-formed.
It also distinguishes the training-time syntax reward from the evaluation-time Syntax Acc. metric: the former gives partial credit over local syntactic units, whereas the latter checks whether the complete generated program is syntactically valid.

\section{Error-Type Analysis}

To complement the aggregate benchmark scores in the main paper, we categorize each generated program by its final execution outcome.
The five categories are mutually exclusive: a sample is assigned to the first applicable category among syntax error, runtime error, timeout, wrong answer, and accepted.
Consequently, every row in Table~\ref{tab:error_distribution} sums to 100\%. For each seed, the error-type percentages are computed separately for each benchmark and then macro-averaged across the six benchmarks, assigning equal weight to each benchmark. The reported values are subsequently averaged over three seeds. Under this aggregation, the Accepted proportion corresponds to the average Pass@1 reported in the main paper.

Compared with the base model, DHRCL reduces syntax errors from 8.0\% to 4.8\%, runtime errors from 9.7\% to 5.9\%, and timeouts from 4.8\% to 2.9\%.
Relative to VeRPO, DHRCL also lowers all three invalid-program categories and increases the accepted proportion from 49.2\% to 50.3\%.
Executable-but-incorrect programs remain the dominant residual failure mode after syntax and execution failures are reduced.

\section{Detailed Ablation Results}
\label{app:detailed_ablation_results}

Table~\ref{tab:reward_ablation_full} reports the benchmark-level reward-component ablation, while Table~\ref{tab:curriculum_prob_ablation_full} summarizes the curriculum-scheduling and token-credit ablations. Stage Transitions report the median first/second transition
iterations over three seeds.
Together, these results isolate the contributions of reward composition, trend-based stage progression, and stage-aware probability-based token credit redistribution.

\paragraph{Reward components.}
As shown in Table~\ref{tab:reward_ablation_full}, adding syntax and execution feedback improves intermediate program validity, while the full reward obtains the highest average Pass@1 and AST similarity.
The comparison between \textit{Pass + Syntax + Exec} and \textit{Full Reward} associates the AST component with a 0.6-point increase in average Pass@1 and a clearer improvement in AST similarity.
We therefore interpret structural alignment as the primary direct effect and the Pass@1 gain as a possible secondary benefit.

\paragraph{Curriculum and token credit.}
Table~\ref{tab:curriculum_prob_ablation_full} shows that trend-based stage progression outperforms both fixed reward aggregation and predefined stage switching.
Among token-credit variants, the complete consolidation-to-refinement ordering achieves the highest average Pass@1, the lowest DGR, and the smallest normalized training cost.
The reversed variant performs worse than either the early-only or late-only variants, supporting the importance of the proposed stage ordering.

\FloatBarrier

\bibliography{aaai2027}

@article{survey01,
  author       = {Junqiao Wang and
                  Zeng Zhang and
                  Yangfan He and
                  Yuyang Song and
                  Tianyu Shi and
                  Yuchen Li and
                  Hengyuan Xu and
                  Kunyu Wu and
                  Guangwu Qian and
                  Qiuwu Chen and
                  Lewei He},
  title        = {Enhancing Code LLMs with Reinforcement Learning in Code Generation:
                  {A} Survey},
  journal      = {CoRR},
  volume       = {abs/2412.20367},
  year         = {2024},
  url          = {https://doi.org/10.48550/arXiv.2412.20367},
  doi          = {10.48550/ARXIV.2412.20367},
  eprinttype   = {arXiv},
  eprint       = {2412.20367},
  bibsource    = {dblp computer science bibliography, https://dblp.org}
}

@article{survey02,
  author       = {Juyong Jiang and
                  Fan Wang and
                  Jiasi Shen and
                  Sungju Kim and
                  Sung Hun Kim},
  title        = {A Survey on Large Language Models for Code Generation},
  journal      = {{ACM} Trans. Softw. Eng. Methodol.},
  volume       = {35},
  number       = {2},
  pages        = {58:1--58:72},
  year         = {2026},
  url          = {https://doi.org/10.1145/3747588},
  doi          = {10.1145/3747588},
  bibsource    = {dblp computer science bibliography, https://dblp.org}
}

@inproceedings{coderl,
  author       = {Hung Le and
                  Yue Wang and
                  Akhilesh Deepak Gotmare and
                  Silvio Savarese and
                  Steven Chu{-}Hong Hoi},
  editor       = {Sanmi Koyejo and
                  S. Mohamed and
                  A. Agarwal and
                  Danielle Belgrave and
                  K. Cho and
                  A. Oh},
  title        = {CodeRL: Mastering Code Generation through Pretrained Models and Deep
                  Reinforcement Learning},
  booktitle    = {Advances in Neural Information Processing Systems 35: Annual Conference
                  on Neural Information Processing Systems 2022, NeurIPS 2022, New Orleans,
                  LA, USA, November 28 - December 9, 2022},
  year         = {2022},
  url          = {http://papers.nips.cc/paper\_files/paper/2022/hash/8636419dea1aa9fbd25fc4248e702da4-Abstract-Conference.html},
  bibsource    = {dblp computer science bibliography, https://dblp.org}
}

@article{ppocoder,
  author       = {Parshin Shojaee and
                  Aneesh Jain and
                  Sindhu Tipirneni and
                  Chandan K. Reddy},
  title        = {Execution-based Code Generation using Deep Reinforcement Learning},
  journal      = {Trans. Mach. Learn. Res.},
  volume       = {2023},
  year         = {2023},
  url          = {https://openreview.net/forum?id=0XBuaxqEcG},
  bibsource    = {dblp computer science bibliography, https://dblp.org}
}

@article{RLTF,
  author       = {Jiate Liu and
                  Yiqin Zhu and
                  Kaiwen Xiao and
                  Qiang Fu and
                  Xiao Han and
                  Wei Yang and
                  Deheng Ye},
  title        = {{RLTF:} Reinforcement Learning from Unit Test Feedback},
  journal      = {Trans. Mach. Learn. Res.},
  volume       = {2023},
  year         = {2023},
  url          = {https://openreview.net/forum?id=hjYmsV6nXZ},
  bibsource    = {dblp computer science bibliography, https://dblp.org}
}

@article{dsmath,
  author       = {Zhihong Shao and
                  Peiyi Wang and
                  Qihao Zhu and
                  Runxin Xu and
                  Junxiao Song and
                  Mingchuan Zhang and
                  Y. K. Li and
                  Y. Wu and
                  Daya Guo},
  title        = {DeepSeekMath: Pushing the Limits of Mathematical Reasoning in Open
                  Language Models},
  journal      = {CoRR},
  volume       = {abs/2402.03300},
  year         = {2024},
  url          = {https://doi.org/10.48550/arXiv.2402.03300},
  doi          = {10.48550/ARXIV.2402.03300},
  eprinttype   = {arXiv},
  eprint       = {2402.03300},
  bibsource    = {dblp computer science bibliography, https://dblp.org}
}

@article{dsr1,
  author       = {DeepSeek{-}AI},
  title        = {DeepSeek-R1: Incentivizing Reasoning Capability in LLMs via Reinforcement
                  Learning},
  journal      = {CoRR},
  volume       = {abs/2501.12948},
  year         = {2025},
  url          = {https://doi.org/10.48550/arXiv.2501.12948},
  doi          = {10.48550/ARXIV.2501.12948},
  eprinttype   = {arXiv},
  eprint       = {2501.12948},
  bibsource    = {dblp computer science bibliography, https://dblp.org}
}

@article{ppo,
  author       = {John Schulman and
                  Filip Wolski and
                  Prafulla Dhariwal and
                  Alec Radford and
                  Oleg Klimov},
  title        = {Proximal Policy Optimization Algorithms},
  journal      = {CoRR},
  volume       = {abs/1707.06347},
  year         = {2017},
  url          = {http://arxiv.org/abs/1707.06347},
  eprinttype   = {arXiv},
  eprint       = {1707.06347},
  bibsource    = {dblp computer science bibliography, https://dblp.org}
}

@article{reinforce,
  author       = {Ronald J. Williams},
  title        = {Simple Statistical Gradient-Following Algorithms for Connectionist
                  Reinforcement Learning},
  journal      = {Mach. Learn.},
  volume       = {8},
  pages        = {229--256},
  year         = {1992},
  url          = {https://doi.org/10.1007/BF00992696},
  doi          = {10.1007/BF00992696},
  bibsource    = {dblp computer science bibliography, https://dblp.org}
}

@article{dapo,
  author       = {Qiying Yu and
                  Zheng Zhang and
                  Ruofei Zhu and
                  Yufeng Yuan and
                  Xiaochen Zuo and
                  Yu Yue and
                  Tiantian Fan and
                  Gaohong Liu and
                  Lingjun Liu and
                  Xin Liu and
                  Haibin Lin and
                  Zhiqi Lin and
                  Bole Ma and
                  Guangming Sheng and
                  Yuxuan Tong and
                  Chi Zhang and
                  Mofan Zhang and
                  Wang Zhang and
                  Hang Zhu and
                  Jinhua Zhu and
                  Jiaze Chen and
                  Jiangjie Chen and
                  Chengyi Wang and
                  Hongli Yu and
                  Weinan Dai and
                  Yuxuan Song and
                  Xiangpeng Wei and
                  Hao Zhou and
                  Jingjing Liu and
                  Wei{-}Ying Ma and
                  Ya{-}Qin Zhang and
                  Lin Yan and
                  Mu Qiao and
                  Yonghui Wu and
                  Mingxuan Wang},
  title        = {{DAPO:} An Open-Source {LLM} Reinforcement Learning System at Scale},
  journal      = {CoRR},
  volume       = {abs/2503.14476},
  year         = {2025},
  url          = {https://doi.org/10.48550/arXiv.2503.14476},
  doi          = {10.48550/ARXIV.2503.14476},
  eprinttype   = {arXiv},
  eprint       = {2503.14476},
  bibsource    = {dblp computer science bibliography, https://dblp.org}
}

@inproceedings{RLEF,
  author       = {Jonas Gehring and
                  Kunhao Zheng and
                  Jade Copet and
                  Vegard Mella and
                  Taco Cohen and
                  Gabriel Synnaeve},
  editor       = {Aarti Singh and
                  Maryam Fazel and
                  Daniel Hsu and
                  Simon Lacoste{-}Julien and
                  Felix Berkenkamp and
                  Tegan Maharaj and
                  Kiri Wagstaff and
                  Jerry Zhu},
  title        = {{RLEF:} Grounding Code LLMs in Execution Feedback with Reinforcement
                  Learning},
  booktitle    = {Forty-second International Conference on Machine Learning, {ICML}
                  2025, Vancouver, BC, Canada, July 13-19, 2025},
  series       = {Proceedings of Machine Learning Research},
  publisher    = {{PMLR} / OpenReview.net},
  year         = {2025},
  url          = {https://proceedings.mlr.press/v267/gehring25a.html},
  bibsource    = {dblp computer science bibliography, https://dblp.org}
}

@inproceedings{dpo,
  author       = {Rafael Rafailov and
                  Archit Sharma and
                  Eric Mitchell and
                  Christopher D. Manning and
                  Stefano Ermon and
                  Chelsea Finn},
  editor       = {Alice Oh and
                  Tristan Naumann and
                  Amir Globerson and
                  Kate Saenko and
                  Moritz Hardt and
                  Sergey Levine},
  title        = {Direct Preference Optimization: Your Language Model is Secretly a
                  Reward Model},
  booktitle    = {Advances in Neural Information Processing Systems 36: Annual Conference
                  on Neural Information Processing Systems 2023, NeurIPS 2023, New Orleans,
                  LA, USA, December 10 - 16, 2023},
  year         = {2023},
  url          = {http://papers.nips.cc/paper\_files/paper/2023/hash/a85b405ed65c6477a4fe8302b5e06ce7-Abstract-Conference.html},
  bibsource    = {dblp computer science bibliography, https://dblp.org}
}

@inproceedings{focuseddpo,
  author       = {Kechi Zhang and
                  Ge Li and
                  Jia Li and
                  Yihong Dong and
                  Zhi Jin},
  editor       = {Wanxiang Che and
                  Joyce Nabende and
                  Ekaterina Shutova and
                  Mohammad Taher Pilehvar},
  title        = {Focused-DPO: Enhancing Code Generation Through Focused Preference
                  Optimization on Error-Prone Points},
  booktitle    = {Findings of the Association for Computational Linguistics, {ACL} 2025,
                  Vienna, Austria, July 27 - August 1, 2025},
  series       = {Findings of {ACL}},
  pages        = {9578--9591},
  publisher    = {Association for Computational Linguistics},
  year         = {2025},
  url          = {https://aclanthology.org/2025.findings-acl.498/},
  bibsource    = {dblp computer science bibliography, https://dblp.org}
}

@article{gspo,
  author       = {Chujie Zheng and
                  Shixuan Liu and
                  Mingze Li and
                  Xiong{-}Hui Chen and
                  Bowen Yu and
                  Chang Gao and
                  Kai Dang and
                  Yuqiong Liu and
                  Rui Men and
                  An Yang and
                  Jingren Zhou and
                  Junyang Lin},
  title        = {Group Sequence Policy Optimization},
  journal      = {CoRR},
  volume       = {abs/2507.18071},
  year         = {2025},
  url          = {https://doi.org/10.48550/arXiv.2507.18071},
  doi          = {10.48550/ARXIV.2507.18071},
  eprinttype   = {arXiv},
  eprint       = {2507.18071},
  bibsource    = {dblp computer science bibliography, https://dblp.org}
}

@inproceedings{RLHWGen,
  author       = {Yifang Zhao and
                  Weimin Fu and
                  Shijie Li and
                  Yixiang Hu and
                  Xiaolong Guo and
                  Yier Jin},
  title        = {Enhancing {LLM} Performance on Hardware Design Generation Task via
                  Reinforcement Learning},
  booktitle    = {{IEEE} International Symposium on Circuits and Systems, {ISCAS} 2025,
                  London, United Kingdom, May 25-28, 2025},
  pages        = {1--5},
  publisher    = {{IEEE}},
  year         = {2025},
  url          = {https://doi.org/10.1109/ISCAS56072.2025.11043832},
  doi          = {10.1109/ISCAS56072.2025.11043832},
  bibsource    = {dblp computer science bibliography, https://dblp.org}
}

@article{VeRPO,
  author       = {Longwen Wang and
                  Xuan'er Wu and
                  Xiaohui Hu and
                  Yirui Liu and
                  Yuankai Fan and
                  Kaidong Yu and
                  Qizhen Weng and
                  Wei Xi and
                  Xuelong Li},
  title        = {VeRPO: Verifiable Dense Reward Policy Optimization for Code Generation},
  journal      = {CoRR},
  volume       = {abs/2601.03525},
  year         = {2026},
  url          = {https://doi.org/10.48550/arXiv.2601.03525},
  doi          = {10.48550/ARXIV.2601.03525},
  eprinttype   = {arXiv},
  eprint       = {2601.03525},
  bibsource    = {dblp computer science bibliography, https://dblp.org}
}

@inproceedings{ReCode,
  author       = {Haoze Wu and
                  Yunzhi Yao and
                  Wenhao Yu and
                  Ningyu Zhang},
  editor       = {Sven Koenig and
                  Chad Jenkins and
                  Matthew E. Taylor},
  title        = {ReCode: Updating Code {API} Knowledge with Reinforcement Learning},
  booktitle    = {Fortieth {AAAI} Conference on Artificial Intelligence, Thirty-Eighth
                  Conference on Innovative Applications of Artificial Intelligence,
                  Sixteenth Symposium on Educational Advances in Artificial Intelligence,
                  {AAAI} 2026, Singapore, January 20-27, 2026},
  pages        = {33908--33916},
  publisher    = {{AAAI} Press},
  year         = {2026},
  url          = {https://doi.org/10.1609/aaai.v40i40.40683},
  doi          = {10.1609/AAAI.V40I40.40683},
  bibsource    = {dblp computer science bibliography, https://dblp.org}
}

@inproceedings{AceCoder,
  author       = {Huaye Zeng and
                  Dongfu Jiang and
                  Haozhe Wang and
                  Ping Nie and
                  Xiaotong Chen and
                  Wenhu Chen},
  editor       = {Wanxiang Che and
                  Joyce Nabende and
                  Ekaterina Shutova and
                  Mohammad Taher Pilehvar},
  title        = {{ACECODER:} Acing Coder {RL} via Automated Test-Case Synthesis},
  booktitle    = {Proceedings of the 63rd Annual Meeting of the Association for Computational
                  Linguistics (Volume 1: Long Papers), {ACL} 2025, Vienna, Austria,
                  July 27 - August 1, 2025},
  pages        = {12023--12040},
  publisher    = {Association for Computational Linguistics},
  year         = {2025},
  url          = {https://aclanthology.org/2025.acl-long.587/},
  bibsource    = {dblp computer science bibliography, https://dblp.org}
}

@article{CodePRM,
  author       = {Ning Dai and
                  Zheng Wu and
                  Renjie Zheng and
                  Ziyun Wei and
                  Wenlei Shi and
                  Xing Jin and
                  Guanlin Liu and
                  Chen Dun and
                  Liang Huang and
                  Lin Yan},
  title        = {Process Supervision-Guided Policy Optimization for Code Generation},
  journal      = {CoRR},
  volume       = {abs/2410.17621},
  year         = {2024},
  url          = {https://doi.org/10.48550/arXiv.2410.17621},
  doi          = {10.48550/ARXIV.2410.17621},
  eprinttype   = {arXiv},
  eprint       = {2410.17621},
  bibsource    = {dblp computer science bibliography, https://dblp.org}
}

@inproceedings{Process-SupervisedRL,
  author       = {Yufan Ye and
                  Ting Zhang and
                  Wenbin Jiang and
                  Hua Huang},
  editor       = {Christos Christodoulopoulos and
                  Tanmoy Chakraborty and
                  Carolyn Rose and
                  Violet Peng},
  title        = {Process-Supervised Reinforcement Learning for Code Generation},
  booktitle    = {Proceedings of the 2025 Conference on Empirical Methods in Natural
                  Language Processing, {EMNLP} 2025, Suzhou, China, November 4-9, 2025},
  pages        = {14213--14226},
  publisher    = {Association for Computational Linguistics},
  year         = {2025},
  url          = {https://doi.org/10.18653/v1/2025.emnlp-main.719},
  doi          = {10.18653/V1/2025.EMNLP-MAIN.719},
  bibsource    = {dblp computer science bibliography, https://dblp.org}
}

@inproceedings{Target-DPO,
  author       = {Jie Wu and
                  Haoling Li and
                  Xin Zhang and
                  Xiao Liu and
                  Yangyu Huang and
                  Jianwen Luo and
                  Yizhen Zhang and
                  Zuchao Li and
                  Ruihang Chu and
                  Yujiu Yang and
                  Scarlett Li},
  editor       = {Christos Christodoulopoulos and
                  Tanmoy Chakraborty and
                  Carolyn Rose and
                  Violet Peng},
  title        = {Teaching Your Models to Understand Code via Focal Preference Alignment},
  booktitle    = {Proceedings of the 2025 Conference on Empirical Methods in Natural
                  Language Processing, {EMNLP} 2025, Suzhou, China, November 4-9, 2025},
  pages        = {14003--14023},
  publisher    = {Association for Computational Linguistics},
  year         = {2025},
  url          = {https://doi.org/10.18653/v1/2025.emnlp-main.707},
  doi          = {10.18653/V1/2025.EMNLP-MAIN.707},
  bibsource    = {dblp computer science bibliography, https://dblp.org}
}

@article{Code-A1,
  author       = {Aozhe Wang and
                  Yuchen Yan and
                  Nan Zhou and
                  Zhengxi Lu and
                  Weiming Lu and
                  Jun Xiao and
                  Yueting Zhuang and
                  Yongliang Shen},
  title        = {Code-A1: Adversarial Evolving of Code {LLM} and Test {LLM} via Reinforcement
                  Learning},
  journal      = {CoRR},
  volume       = {abs/2603.15611},
  year         = {2026},
  url          = {https://doi.org/10.48550/arXiv.2603.15611},
  doi          = {10.48550/ARXIV.2603.15611},
  eprinttype   = {arXiv},
  eprint       = {2603.15611},
  bibsource    = {dblp computer science bibliography, https://dblp.org}
}

@inproceedings{CodeDPO,
  author       = {Kechi Zhang and
                  Ge Li and
                  Yihong Dong and
                  Jingjing Xu and
                  Jun Zhang and
                  Jing Su and
                  Yongfei Liu and
                  Zhi Jin},
  editor       = {Wanxiang Che and
                  Joyce Nabende and
                  Ekaterina Shutova and
                  Mohammad Taher Pilehvar},
  title        = {CodeDPO: Aligning Code Models with Self Generated and Verified Source
                  Code},
  booktitle    = {Proceedings of the 63rd Annual Meeting of the Association for Computational
                  Linguistics (Volume 1: Long Papers), {ACL} 2025, Vienna, Austria,
                  July 27 - August 1, 2025},
  pages        = {15854--15871},
  publisher    = {Association for Computational Linguistics},
  year         = {2025},
  url          = {https://aclanthology.org/2025.acl-long.771/},
  bibsource    = {dblp computer science bibliography, https://dblp.org}
}

@article{PLUM,
  author       = {Dylan Zhang and
                  Shizhe Diao and
                  Xueyan Zou and
                  Hao Peng},
  title        = {{PLUM:} Preference Learning Plus Test Cases Yields Better Code Language
                  Models},
  journal      = {CoRR},
  volume       = {abs/2406.06887},
  year         = {2024},
  url          = {https://doi.org/10.48550/arXiv.2406.06887},
  doi          = {10.48550/ARXIV.2406.06887},
  eprinttype   = {arXiv},
  eprint       = {2406.06887},
  bibsource    = {dblp computer science bibliography, https://dblp.org}
}

@article{StepCoder,
  author       = {Shihan Dou and
                  Yan Liu and
                  Haoxiang Jia and
                  Limao Xiong and
                  Enyu Zhou and
                  Wei Shen and
                  Junjie Shan and
                  Caishuang Huang and
                  Xiao Wang and
                  Xiaoran Fan and
                  Zhiheng Xi and
                  Yuhao Zhou and
                  Tao Ji and
                  Rui Zheng and
                  Qi Zhang and
                  Xuanjing Huang and
                  Tao Gui},
  title        = {StepCoder: Improve Code Generation with Reinforcement Learning from
                  Compiler Feedback},
  journal      = {CoRR},
  volume       = {abs/2402.01391},
  year         = {2024},
  url          = {https://doi.org/10.48550/arXiv.2402.01391},
  doi          = {10.48550/ARXIV.2402.01391},
  eprinttype   = {arXiv},
  eprint       = {2402.01391},
  bibsource    = {dblp computer science bibliography, https://dblp.org}
}

@inproceedings{Alignment,
  author       = {Houxing Ren and
                  Zimu Lu and
                  Weikang Shi and
                  Haotian Hou and
                  Yunqiao Yang and
                  Ke Wang and
                  Aojun Zhou and
                  Junting Pan and
                  Mingjie Zhan and
                  Hongsheng Li},
  editor       = {Christos Christodoulopoulos and
                  Tanmoy Chakraborty and
                  Carolyn Rose and
                  Violet Peng},
  title        = {Alignment with Fill-In-the-Middle for Enhancing Code Generation},
  booktitle    = {Proceedings of the 2025 Conference on Empirical Methods in Natural
                  Language Processing, {EMNLP} 2025, Suzhou, China, November 4-9, 2025},
  pages        = {8304--8320},
  publisher    = {Association for Computational Linguistics},
  year         = {2025},
  url          = {https://doi.org/10.18653/v1/2025.emnlp-main.419},
  doi          = {10.18653/V1/2025.EMNLP-MAIN.419},
  bibsource    = {dblp computer science bibliography, https://dblp.org}
}

@inproceedings{small,
  author       = {Marwa Na{\"{\i}}r and
                  Kamel Mohammed Yamani and
                  Lynda Said L'Hadj and
                  Riyadh Baghdadi},
  editor       = {Xiyan Fu and
                  Eve Fleisig},
  title        = {Curriculum Learning for Small Code Language Models},
  booktitle    = {Proceedings of the 62nd Annual Meeting of the Association for Computational
                  Linguistics (Volume 4: Student Research Workshop), {ACL} 2024, Bangkok,
                  Thailand, August 11-16, 2024},
  pages        = {531--542},
  publisher    = {Association for Computational Linguistics},
  year         = {2024},
  url          = {https://doi.org/10.18653/v1/2024.acl-srw.44},
  doi          = {10.18653/V1/2024.ACL-SRW.44},
  bibsource    = {dblp computer science bibliography, https://dblp.org}
}

@article{TAROT,
  author       = {Chansung Park and
                  Juyong Jiang and
                  Fan Wang and
                  Sayak Paul and
                  Jiasi Shen and
                  Jing Tang and
                  Jianguo Li},
  title        = {{TAROT:} Test-driven and Capability-adaptive Curriculum Reinforcement
                  Fine-tuning for Code Generation with Large Language Models},
  journal      = {CoRR},
  volume       = {abs/2602.15449},
  year         = {2026},
  url          = {https://doi.org/10.48550/arXiv.2602.15449},
  doi          = {10.48550/ARXIV.2602.15449},
  eprinttype   = {arXiv},
  eprint       = {2602.15449},
  bibsource    = {dblp computer science bibliography, https://dblp.org}
}

@article{codex,
  author       = {Mark Chen and
                  Jerry Tworek and
                  Heewoo Jun and
                  Qiming Yuan and
                  Henrique Pond{\'{e}} de Oliveira Pinto and
                  Jared Kaplan and
                  Harri Edwards and
                  Yuri Burda and
                  Nicholas Joseph and
                  Greg Brockman and
                  Alex Ray and
                  Raul Puri and
                  Gretchen Krueger and
                  Michael Petrov and
                  Heidy Khlaaf and
                  Girish Sastry and
                  Pamela Mishkin and
                  Brooke Chan and
                  Scott Gray and
                  Nick Ryder and
                  Mikhail Pavlov and
                  Alethea Power and
                  Lukasz Kaiser and
                  Mohammad Bavarian and
                  Clemens Winter and
                  Philippe Tillet and
                  Felipe Petroski Such and
                  Dave Cummings and
                  Matthias Plappert and
                  Fotios Chantzis and
                  Elizabeth Barnes and
                  Ariel Herbert{-}Voss and
                  William Hebgen Guss and
                  Alex Nichol and
                  Alex Paino and
                  Nikolas Tezak and
                  Jie Tang and
                  Igor Babuschkin and
                  Suchir Balaji and
                  Shantanu Jain and
                  William Saunders and
                  Christopher Hesse and
                  Andrew N. Carr and
                  Jan Leike and
                  Joshua Achiam and
                  Vedant Misra and
                  Evan Morikawa and
                  Alec Radford and
                  Matthew Knight and
                  Miles Brundage and
                  Mira Murati and
                  Katie Mayer and
                  Peter Welinder and
                  Bob McGrew and
                  Dario Amodei and
                  Sam McCandlish and
                  Ilya Sutskever and
                  Wojciech Zaremba},
  title        = {Evaluating Large Language Models Trained on Code},
  journal      = {CoRR},
  volume       = {abs/2107.03374},
  year         = {2021},
  url          = {https://arxiv.org/abs/2107.03374},
  eprinttype   = {arXiv},
  eprint       = {2107.03374},
  bibsource    = {dblp computer science bibliography, https://dblp.org}
}

@article{codellama,
  author       = {Baptiste Rozi{\`{e}}re and
                  Jonas Gehring and
                  Fabian Gloeckle and
                  Sten Sootla and
                  Itai Gat and
                  Xiaoqing Ellen Tan and
                  Yossi Adi and
                  Jingyu Liu and
                  Tal Remez and
                  J{\'{e}}r{\'{e}}my Rapin and
                  Artyom Kozhevnikov and
                  Ivan Evtimov and
                  Joanna Bitton and
                  Manish Bhatt and
                  Cristian Canton{-}Ferrer and
                  Aaron Grattafiori and
                  Wenhan Xiong and
                  Alexandre D{\'{e}}fossez and
                  Jade Copet and
                  Faisal Azhar and
                  Hugo Touvron and
                  Louis Martin and
                  Nicolas Usunier and
                  Thomas Scialom and
                  Gabriel Synnaeve},
  title        = {Code Llama: Open Foundation Models for Code},
  journal      = {CoRR},
  volume       = {abs/2308.12950},
  year         = {2023},
  url          = {https://doi.org/10.48550/arXiv.2308.12950},
  doi          = {10.48550/ARXIV.2308.12950},
  eprinttype   = {arXiv},
  eprint       = {2308.12950},
  bibsource    = {dblp computer science bibliography, https://dblp.org}
}

@inproceedings{bigcodebench,
  author       = {Terry Yue Zhuo and
                  Minh Chien Vu and
                  Jenny Chim and
                  Han Hu and
                  Wenhao Yu and
                  Ratnadira Widyasari and
                  Imam Nur Bani Yusuf and
                  Haolan Zhan and
                  Junda He and
                  Indraneil Paul and
                  Simon Brunner and
                  Chen Gong and
                  James Hoang and
                  Armel Randy Zebaze and
                  Xiaoheng Hong and
                  Wen{-}Ding Li and
                  Jean Kaddour and
                  Ming Xu and
                  Zhihan Zhang and
                  Prateek Yadav and
                  et al.},
  title        = {BigCodeBench: Benchmarking Code Generation with Diverse Function Calls
                  and Complex Instructions},
  booktitle    = {The Thirteenth International Conference on Learning Representations,
                  {ICLR} 2025, Singapore, April 24-28, 2025},
  publisher    = {OpenReview.net},
  year         = {2025},
  url          = {https://openreview.net/forum?id=YrycTjllL0},
  bibsource    = {dblp computer science bibliography, https://dblp.org}
}

@inproceedings{odex,
  author       = {Zhiruo Wang and
                  Shuyan Zhou and
                  Daniel Fried and
                  Graham Neubig},
  editor       = {Houda Bouamor and
                  Juan Pino and
                  Kalika Bali},
  title        = {Execution-Based Evaluation for Open-Domain Code Generation},
  booktitle    = {Findings of the Association for Computational Linguistics: {EMNLP}
                  2023, Singapore, December 6-10, 2023},
  series       = {Findings of {ACL}},
  pages        = {1271--1290},
  publisher    = {Association for Computational Linguistics},
  year         = {2023},
  url          = {https://doi.org/10.18653/v1/2023.findings-emnlp.89},
  doi          = {10.18653/V1/2023.FINDINGS-EMNLP.89},
  bibsource    = {dblp computer science bibliography, https://dblp.org}
}

@article{yang2025qwen3,
 author       = {Qwen Team},
  title        = {Qwen3 Technical Report},
  journal      = {CoRR},
  volume       = {abs/2505.09388},
  year         = {2025},
  url          = {https://doi.org/10.48550/arXiv.2505.09388},
  doi          = {10.48550/ARXIV.2505.09388},
  eprinttype   = {arXiv},
  eprint       = {2505.09388},
  bibsource    = {dblp computer science bibliography, https://dblp.org}
}

@article{wang2026kodcode,
author       = {Zhangchen Xu and
                  Yang Liu and
                  Yueqin Yin and
                  Mingyuan Zhou and
                  Radha Poovendran},
  editor       = {Wanxiang Che and
                  Joyce Nabende and
                  Ekaterina Shutova and
                  Mohammad Taher Pilehvar},
  title        = {KodCode: {A} Diverse, Challenging, and Verifiable Synthetic Dataset
                  for Coding},
  booktitle    = {Findings of the Association for Computational Linguistics, {ACL} 2025,
                  Vienna, Austria, July 27 - August 1, 2025},
  series       = {Findings of {ACL}},
  pages        = {6980--7008},
  publisher    = {Association for Computational Linguistics},
  year         = {2025},
  url          = {https://doi.org/10.18653/v1/2025.findings-acl.365},
  doi          = {10.18653/V1/2025.FINDINGS-ACL.365},
  bibsource    = {dblp computer science bibliography, https://dblp.org}
}

@article{liu2023humanevalplus,
author       = {Jiawei Liu and
                  Chunqiu Steven Xia and
                  Yuyao Wang and
                  Lingming Zhang},
  editor       = {Alice Oh and
                  Tristan Naumann and
                  Amir Globerson and
                  Kate Saenko and
                  Moritz Hardt and
                  Sergey Levine},
  title        = {Is Your Code Generated by ChatGPT Really Correct? Rigorous Evaluation
                  of Large Language Models for Code Generation},
  booktitle    = {Advances in Neural Information Processing Systems 36: Annual Conference
                  on Neural Information Processing Systems 2023, NeurIPS 2023, New Orleans,
                  LA, USA, December 10 - 16, 2023},
  year         = {2023},
  url          = {http://papers.nips.cc/paper\_files/paper/2023/hash/43e9d647ccd3e4b7b5baab53f0368686-Abstract-Conference.html},
  bibsource    = {dblp computer science bibliography, https://dblp.org}

}

@article{jain2024livecodebench,
   author       = {Naman Jain and
                  King Han and
                  Alex Gu and
                  Wen{-}Ding Li and
                  Fanjia Yan and
                  Tianjun Zhang and
                  Sida Wang and
                  Armando Solar{-}Lezama and
                  Koushik Sen and
                  Ion Stoica},
  title        = {LiveCodeBench: Holistic and Contamination Free Evaluation of Large
                  Language Models for Code},
  booktitle    = {The Thirteenth International Conference on Learning Representations,
                  {ICLR} 2025, Singapore, April 24-28, 2025},
  publisher    = {OpenReview.net},
  year         = {2025},
  url          = {https://openreview.net/forum?id=chfJJYC3iL},
  bibsource    = {dblp computer science bibliography, https://dblp.org}
}

@article{quan2025codeelo,
 author       = {Shanghaoran Quan and
                  Jiaxi Yang and
                  Bowen Yu and
                  Bo Zheng and
                  Dayiheng Liu and
                  An Yang and
                  Xuancheng Ren and
                  Bofei Gao and
                  Yibo Miao and
                  Yunlong Feng and
                  Zekun Wang and
                  Jian Yang and
                  Zeyu Cui and
                  Yang Fan and
                  Yichang Zhang and
                  Binyuan Hui and
                  Junyang Lin},
  title        = {CodeElo: Benchmarking Competition-level Code Generation of LLMs with
                  Human-comparable Elo Ratings},
  journal      = {CoRR},
  volume       = {abs/2501.01257},
  year         = {2025},
  url          = {https://doi.org/10.48550/arXiv.2501.01257},
  doi          = {10.48550/ARXIV.2501.01257},
  eprinttype   = {arXiv},
  eprint       = {2501.01257},
  bibsource    = {dblp computer science bibliography, https://dblp.org}
}
\end{document}